\documentclass[runningheads]{llncs}

\usepackage{eccv}
\usepackage{eccvabbrv}

\usepackage{graphicx}
\usepackage{booktabs}
\usepackage{multirow}
\usepackage{xcolor}
\usepackage{float}
\usepackage{subcaption}

\usepackage[accsupp]{axessibility}  % Improves PDF readability for those with disabilities.

\usepackage{hyperref}
\hypersetup{colorlinks=true, urlcolor=purple}

\usepackage{orcidlink}

\newcommand{\repeatthanks}{\textsuperscript{\thefootnote}}

\begin{document}

% \title{InstantHDR: Single-forward Gaussian Splatting for High Dynamic Range 3D Reconstruction}
\title{InstantHDR: Single-forward Gaussian Splatting Initialization for HDR 3D Reconstruction}

\titlerunning{Single-forward Gaussian Splatting Initialization for HDR 3D Reconstruction}

\author{Dingqiang Ye\thanks{co-first authors; $^{\dagger}$corresponding author}\inst{1} \and
Jiacong Xu\repeatthanks\inst{1}\and
Jianglu Ping\inst{1}\and \\
Yuxiang Guo\inst{1}\and
Chao Fan\inst{2}\and
Vishal M. Patel$^{\dagger,1}$
}

\authorrunning{D.~Ye, J.~Xu et al.}

% % TODO FINAL: Replace with your institution list.
% \institute{Johns Hopkins University, Baltimore MD 21218, USA \and
% Shenzhen University, Guangdong 518060, China

\institute{Johns Hopkins University, USA \and
Shenzhen University, China \\
\email{\{dye6,jxu155,jping1,yguo87\}@jhu.edu}, \\
\email{chaofan996@szu.edu.cn}, 
\email{vpatel36@jhu.edu}
}

\maketitle

\begin{abstract}
High dynamic range (HDR) novel view synthesis (NVS) aims to reconstruct HDR scenes from multi-exposure low dynamic range (LDR) images.
Existing HDR pipelines heavily rely on known camera poses, well-initialized dense point clouds, and time-consuming per-scene optimization.
Current feed-forward alternatives overlook the HDR problem by assuming exposure-invariant appearance.
To bridge this gap, we propose InstantHDR, a feed-forward network that initializes 3D HDR scenes from uncalibrated multi-exposure LDR collections in a fast single forward pass.
Specifically, we design a geometry-guided appearance modeling for multi-exposure fusion, and a meta-network for generalizable scene-specific tone mapping.
Due to the lack of HDR scene data, we build a pre-training dataset, called HDR-Pretrain, for generalizable feed-forward HDR models, featuring 168 Blender-rendered scenes, diverse lighting types, and multiple camera response functions.
Comprehensive experiments show that our InstantHDR delivers a single-forward HDR initialization at $\sim700\times$ the speed of SoTA optimization-based methods, and reaches comparable quality in real settings after lightweight post-optimization while remaining $\sim20\times$ faster.
% All code, models, and datasets will be released after the review process.
All code, models, and datasets: \url{https://github.com/Bugjudger/InstantHDR}.

\end{abstract}

\section{Introduction}
\label{sec:intro}
% 1. HDR任务是什么；
% 2. 以前的HDR方法怎么做，主要是在设计xxx，根本问题是时间久，不可泛化；feed-forward开始流行，但缺少相关设计。
% 3. 面临的挑战：1. 不同视角的多曝光融合。直接融合明亮的图片a和昏暗的图片b，会导致外观重建不一致。2. 每个场景的相机映射函数不同，如索尼 尼康等等都有各自的色彩偏好。3. 缺少相关HDR数据集进行大规模预训练。
% 4. InstantHDR：xxxxx
% 5. 实验如何，contribution如何

% 1. HDR任务是什么；
High dynamic range (HDR) novel view synthesis (NVS) aims to reconstruct HDR scenes from multi-view low dynamic range (LDR) images captured at varying exposures.
Unlike typical low dynamic range (from 0 to 255) imaging, which often suffers from detail loss in extreme lighting and color distortion due to sensor limitations, HDR captures a broader spectrum of luminance (from 0 to +$\infty$).
Through integrating advanced frameworks like NeRF~\cite{mildenhall2021nerf} or 3D Gaussian Splatting~\cite{3dgs} with a tone mapper to model the camera response function (CRF), HDR task enables the re-rendering of photo-realistic novel views with controllable exposure.
Its ability to faithfully represent real-world light and shadow makes it indispensable for multiple applications such as autonomous driving~\cite{yang2020surfelgan,huang2023neural,wang2023sparsenerf,tancik2022block}, digital humans~\cite{liu2021neural,hu2021egorenderer,zheng2023ilsh,zheng2022structured}, and immersive image editing~\cite{liu2021editing,sun2022ide,yuan2022nerf,kobayashi2022decomposing}.

\begin{figure}[t]
\centering
\includegraphics[width=0.95\linewidth]{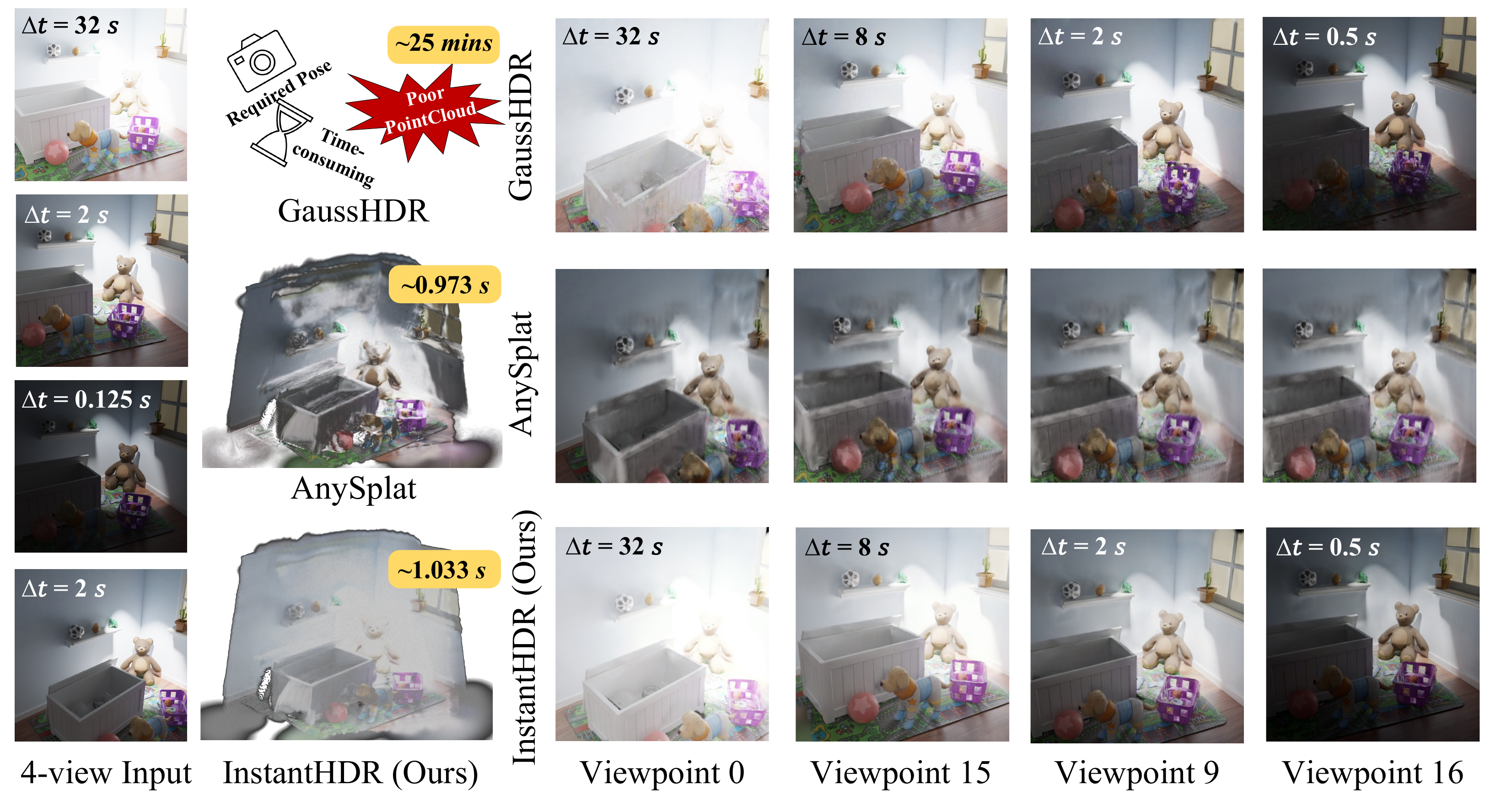}
\caption{
Comparisons of reconstruction time (yellow boxes), scenes (left) and rendered views (right) between the GaussHDR~\cite{liu2025gausshdr} (top), original AnySplat~\cite{jiang2025anysplat} (middle) and our InstantHDR (bottom).
(i) GaussHDR~\cite{liu2025gausshdr} spends expensive $25~mins$ and produces tearing artifacts, as its initial point clouds collapse under the sparse-view inputs.
(ii) AnySplat~\cite{jiang2025anysplat} naively fuses multi-exposure inputs, causing ghosting artifacts and lacking exposure control.
(iii) Our InstantHDR initializes 3D-consistent HDR scenes in few seconds and renders clean LDR images with controllable exposure time.
}
\label{fig:1_compare}
\end{figure}

% 2. 以前的HDR方法怎么做，主要是在设计xxx，根本问题是时间久，不可泛化；feed-forward开始流行，但缺少相关设计。
Existing HDR NVS methods~\cite{huang2022hdr,cai2024hdr,liu2025gausshdr} are predominantly optimization-based.
However, this paradigm is costly and generalizes poorly: it relies on precisely calibrated camera poses and dense multi-view inputs for SfM-based point cloud initialization, followed by time-consuming per-scene densification and optimization.
For example, as shown in Fig.~\ref{fig:1_compare}, GaussHDR~\cite{liu2025gausshdr} struggles to reconstruct scenes from only four LDR views, since exposure-induced appearance inconsistencies degrade the reliability of SfM-based point cloud initialization in sparse-view settings. Furthermore, their heavy computational overhead and strong data dependency limit practical deployment in real-time scenarios.

% 3. 面临的挑战：1. 不同视角的多曝光融合。直接融合明亮的图片a和昏暗的图片b，会导致外观重建不一致。2. 如何在曝光不同的图片中找到几何对应区域？ 3. 每个场景的相机映射函数不同，如索尼 尼康等等都有各自的色彩偏好。4. 缺少相关HDR数据集进行大规模预训练。
Recently, 3D feed-forward  models~\cite{wang2025vggt,jiang2025anysplat} have revolutionized scene reconstruction by inferring geometry in seconds, achieving impressive speed and better generalization than optimization-based methods. Integrating this paradigm into the HDR NVS tasks could boost model generalizability and inference speed.
However, directly applying the original feed-forward models to HDR reconstruction may encounter the following issues. 
(a) \textbf{Exposure-induced Appearance Inconsistency}: 
Naive fusion leads to severe ghosting — as shown in Fig.~\ref{fig:1_compare}. The same white wall appears bright at $\Delta=32s$ but nearly black at $\Delta=0.125s$, causing visible artifacts in AnySplat~\cite{jiang2025anysplat}.
(b) \textbf{Pixel-level Geometric Alignment}: 
Establishing accurate pixel-level geometric correspondences remains a non-trivial task under large brightness variations.
(c) \textbf{Camera Response Functions Inconsistency}: 
In real world, different camera and software apply distinct color transformations (\textit{e.g.}, AgX, Filmic), making it difficult to learn a unified tone mapping operator.
(d) \textbf{HDR Data Scarcity}: Current publicly available HDR datasets~\cite{huang2022hdr,jun2022hdr} are insufficient (as shown in Tab.~\ref{tab:hdr_datasets}) to support the robust large-scale pre-training required for feed-forward models.

%To address these issues, we propose a novel feed-forward 3D reconstruction method, namely InstantHDR, for HDR novel view synthesis.
%InstantHDR consists of two core designs.
%First, a Geometry-guided Appearance Modeling Module that normalizes multi-exposure LDR inputs into a unified exposure space and leverages geo-attention maps from the geometry encoder to fuse patch-level irradiance features. 
%Fine-grained texture details are recovered by injecting Difference of Gaussians (DoG) high-frequency cues, lifting the representation to pixel-level irradiance before predicting HDR 3D Gaussians. 
%Second, a Meta Net that takes the HDR Gaussians, LDR images, and context exposure times as input to predict scene-specific tonemapper parameters, enabling generalizable HDR-to-LDR rendering with controllable exposure time $\Delta t$.
%The entire pipeline operates in a single forward pass without per-scene optimization, significantly improving reconstruction speed compared to recent optimization-based methods.
%With an optional fast post-optimization, InstantHDR presents superior synthesis quality in sparse-view settings and comparable performance in dense-view scenarios.

To address these challenges, we propose \textbf{InstantHDR}, a novel feed-forward 3D reconstruction framework for HDR novel view synthesis. InstantHDR comprises two key components. First, a \textit{geometry-guided appearance modeling module} normalizes multi-exposure LDR inputs into a unified exposure space and utilizes geo-attention from the geometry encoder to fuse patch-level irradiance features. Fine-grained texture details are further recovered by incorporating Difference of Gaussians (DoG) high-frequency cues, lifting the representation to pixel-level irradiance before predicting HDR 3D Gaussians. 
Second, a \textit{MetaNet} takes the predicted HDR Gaussians, LDR images, and exposure times as input to estimate scene-specific tonemapper parameters, enabling generalizable HDR-to-LDR rendering with controllable exposure time $\Delta t$.
This fast single forward pass yields a strong HDR initialization; a lightweight post-optimization stage then refines it to quality comparable to SoTA, while together remaining orders of magnitude faster than existing fully optimization-based pipelines.

Our contributions can be summarized as follows:

% TODO
% 还有介绍数据集, 这里不需要，在contribution里面提一句也可以

% 下面不用说
%Nevertheless, we observe a challenge for current feed-forward methods in dense-view post-optimization: the coordinate misalignment between predicted Gaussian spaces and ground-truth cameras introduces misguided supervision, limiting the performance upper bound.

\textbf{(i)} 
We propose InstantHDR, the first feed-forward HDR novel view synthesis method. It features a geometry-guided appearance module for exposure-robust multi-view fusion and a meta-network for generalizable tone mapping, enabling 3D HDR initialization from uncalibrated multi-exposure LDRs in seconds.

\textbf{(ii)} 
We build HDR-Pretrain, a large-scale dataset of 168 synthetic indoor scenes to support feed-forward HDR pretraining. 
Experiments show that InstantHDR initializes HDR scenes $\sim$700$\times$ faster than SoTA, reaching competitive quality after lightweight post-optimization while staying $\sim$20$\times$ faster.

\section{Related Works}

\noindent\textbf{High Dynamic Range Imaging.} 
HDR imaging has traditionally merged multiple LDR exposures from a fixed viewpoint~\cite{exposure_fusion} or recovered the camera response function from bracketed LDR sequences~\cite{malik,ward2008high,wu2024fast}, but these methods suffer from ghosting under scene motion.
Subsequent works~\cite{hdr_20,hdr_24,hdr_26,hdr_58,hdr_63} mitigate this via optical-flow-based motion compensation prior to fusion, while learning-based approaches directly learn LDR-to-HDR mappings using CNNs~\cite{hdr_cnn_1,hdr_cnn_2,hdr_cnn_3,hdr_cnn_4} and Transformers~\cite{hdr_trans_1,hdr_trans_2,hdr_trans_3,neural_gaffer,magic,promptfix}.
Others reconstruct HDR from a single LDR image via handcrafted priors in an unsupervised or self-supervised manner~\cite{SelfHDR,2021labeled,SMAE,nazarczuk2022self,fei2023generative,li2024chaos}.
A parallel line operates in the RAW domain, exploiting its higher bit-depth and linear radiometric response—from a single RAW image~\cite{zou2023rawhdr}, by jointly denoising and fusing multi-exposure RAW frames~\cite{liu2023joint,jiang2025learning,lee2025ntire}, or by reconstructing HDR video from alternating-exposure RAW sequences~\cite{shu2024real,xu2024hdrflow}, often coupled with the in-camera ISP~\cite{li2024dualdn}.
However, these methods lack the 3D understanding needed for HDR novel views.

\noindent\textbf{Gaussian Splatting.}
3D Gaussian Splatting (3DGS)~\cite{3dgs} represents scenes as collections of anisotropic Gaussian primitives, enabling real-time rendering through efficient rasterization—offering a significant speed advantage over NeRF-based volumetric ray-marching~\cite{mildenhall2022nerf,jun2022hdr,huang2022hdr}.
This efficiency has driven its adoption across diverse tasks including dynamic scenes~\cite{dynamic3,dynamic1,dynamic2}, SLAM~\cite{slam3,slam4,slam2,slam1}, inverse rendering~\cite{InverseRendering2,InverseRendering3,InverseRendering1}, digital humans~\cite{humangaussian,hugs,gauhuman}, 3D generation~\cite{dreamgaussian,gaussiandreamer,luciddreamer}, and medical imaging~\cite{x_gaussian,r2gs}.
Nevertheless, current HDR extensions~\cite{cai2024hdr,liu2025gausshdr,bolduc2025GaSLight,liu2025mono4dgs,singh24_hdrsplat,gong2025casual3dhdr,li2025sehdr,jin2024lighting,cui2025luminance,zhou2025lita} of 3DGS remain predominantly optimization-based, resulting in expensive per-scene reconstruction times.
Our work aims to fill this gap.

\noindent\textbf{Feed-forward 3D Reconstruction.}
Recent advances pursue end-to-end 3D reconstruction directly from unposed images.
Pioneering works such as DUSt3R~\cite{wang2024dust3r} and MASt3R~\cite{mast3r} replace multi-stage pipelines with a unified model that jointly estimates depth and performs dense scene fusion, and subsequent approaches~\cite{wang20243d, liu2025slam3r, murai2025mast3r, wang2025continuous, wang2025vggt, yang2025fast3r, tang2025mv} cascade transformer blocks to recover camera poses, point trajectories, and scene geometry in a single forward pass.
A parallel line~\cite{jiang2023leap, wang2023pf, hong2024pf3plat, ye2024no, zhang2025flare, smart2024splatt3r, chen2024pref3r} targets novel view synthesis from unposed sparse-view images.
Compared to optimization-based ones, these models offer remarkable speed, strong generalization, and minimal data requirements, yet their potential for HDR reconstruction remains largely unexplored.
We explore this promising direction.

% \noindent\textbf{Feed-forward 3D Reconstruction.}
% Recent advances pursue end-to-end 3D reconstruction directly from unposed images.
% Pioneering works such as DUSt3R~\cite{wang2024dust3r} and MASt3R~\cite{mast3r} replace traditional multi-stage pipelines with a unified model that jointly estimates depth and performs dense scene fusion.
% Subsequent approaches~\cite{wang20243d, liu2025slam3r, murai2025mast3r, wang2025continuous, wang2025vggt, yang2025fast3r, tang2025mv} extend this paradigm by cascading transformer blocks to simultaneously recover camera poses, point trajectories, and scene geometry in a single forward pass.
% A parallel line of research~\cite{jiang2023leap, wang2023pf, hong2024pf3plat, ye2024no, zhang2025flare, smart2024splatt3r, chen2024pref3r} targets novel view synthesis from unposed sparse-view images.
% These feed-forward models offer remarkable reconstruction speed, strong generalization to unseen scenes, and minimal data requirements compared to optimization-based pipelines.
% Their potential for HDR reconstruction remains largely unexplored.
% Our work explores this promising direction.

\section{Method}
\label{sec:method}

\begin{figure}[t]
\centering
\includegraphics[width=\linewidth]{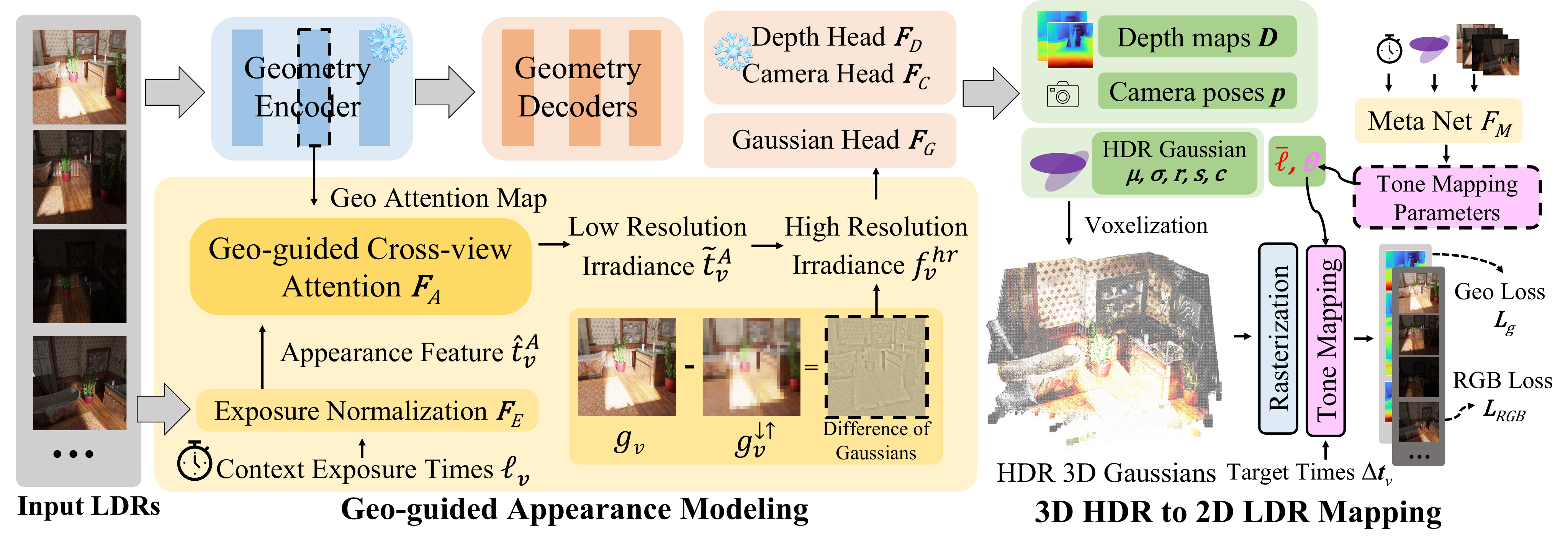}
\caption{
\textbf{Overview of InstantHDR.} 
Given multi-exposure LDR images, the frozen geometry branch estimates depth and camera poses, while the appearance branch normalizes exposures ($F_E$), fuses cross-view irradiance via geometry-guided attention ($F_A$), and recovers pixel-level details via DoG upsampling. 
The Gaussian head $F_G$ combines both branches to produce HDR 3D Gaussians.
The Meta Net $F_M$ predicts tone-mapping parameters for rendering LDR images at controllable exposures.
}
\label{fig:pipeline}
\end{figure}

Given uncalibrated multi-view LDR images captured at varying exposures, InstantHDR initializes HDR 3D Gaussians and renders novel views at any target exposure (Fig.~\ref{fig:pipeline}).
We first formalize the problem in Sec.~\ref{sec:setup}, then detail the pipeline design in Sec.~\ref{sec:pipeline}, including a Geo-guided Appearance Modeling module (Sec.~\ref{sec:appearance}) and a 3D HDR-to-2D LDR Mapping module (Sec.~\ref{sec:tonemapping}), and finally present our training strategies in Sec.~\ref{sec:loss}.

%% =====================================================
\subsection{Problem Setup}
\label{sec:setup}

\paragraph{\textbf{Inputs.}}
Consider $V$ \emph{uncalibrated} views of a single 3D scene, given as context images $\{I_v\}_{v=1}^{V}$, $I_v \in \mathbb{R}^{H \times W \times 3}$, each captured at a known exposure time $\Delta t_v$.
For convenience, we work in log-exposure space and define $\ell_v = \log_2 \Delta t_v$.

\paragraph{\textbf{Outputs.}}
InstantHDR jointly reconstructs the scene geometry and HDR appearance by predicting:

\noindent
\emph{(a)~HDR 3D Gaussians.}
A collection of $G$ anisotropic 3D Gaussians
\begin{equation}
    \bigl\{(\boldsymbol{\mu}_g,\, \sigma_g,\, \boldsymbol{r}_g,\, \boldsymbol{s}_g,\, \boldsymbol{c}_g^{h})\bigr\}_{g=1}^{G},
\end{equation}
where each Gaussian is parameterized by a center position $\boldsymbol{\mu} \in \mathbb{R}^{3}$, an opacity $\sigma \in \mathbb{R}^{+}$, an orientation quaternion $\boldsymbol{r} \in \mathbb{R}^{4}$, an anisotropic scale $\boldsymbol{s} \in \mathbb{R}^{3}$, and an HDR color embedding $\boldsymbol{c}^{h} \in \mathbb{R}^{3 \times (k{+}1)^{2}}$ parameterized as degree-$k$ spherical-harmonic (SH) coefficients that encode \emph{log-radiance}, following~\cite{liu2025gausshdr,cai2024hdr}.

\noindent
\emph{(b)~Camera parameters.}
Per-view parameters $\{p_v \in \mathbb{R}^{9}\}_{v=1}^{V}$, where $p_v$ comprises a focal length, a 3-DoF rotation (axis-angle), a 3-DoF translation, and a 2-DoF principal-point offset.

\noindent
\emph{(c)~Scene-level attributes.}
A mid-exposure anchor $\bar{\ell} = \tfrac{1}{2}(\max_v \ell_v + \min_v \ell_v)$, serving as the reference exposure level, and the parameters $\boldsymbol{\theta}$ of a lightweight tonemapper that approximates the scene-specific camera response function (CRF).

\paragraph{\textbf{Overall mapping.}}
Formally, our model implements:
\begin{equation}
    f_{\boldsymbol{\Theta}}\!:\;
    \{I_v,\, \ell_v\}_{v=1}^{V}
    \;\longmapsto\;
    \Bigl\{(\boldsymbol{\mu}_g,\, \sigma_g,\, \boldsymbol{r}_g,\, \boldsymbol{s}_g,\, \boldsymbol{c}_g^{h})\Bigr\}_{g=1}^{G}
    \;\cup\;
    \{p_v\}_{v=1}^{V}
    \;\cup\;
    \{\bar{\ell},\, \boldsymbol{\theta}\}.
\label{eq:mapping}
\end{equation}
% At test time, given a novel camera pose $p^{*}$ and a target log-exposure $\ell^{*}$, the predicted Gaussians can be rasterized and tonemapped to synthesize a photorealistic LDR image at the desired exposure.

%% =====================================================
\subsection{Pipeline Overview}
\label{sec:pipeline}

As illustrated in Fig.~\ref{fig:pipeline}, our pipeline consists of two branches: a \emph{geometry branch} that estimates scene structure from multi-view images, and an \emph{appearance branch} that reconstructs HDR irradiance from exposure-inconsistent inputs.
Given $V$ uncalibrated multi-exposure LDR images, the geometry branch encodes them into high-dimensional features via a pretrained transformer and decodes depth maps and camera poses.
The appearance branch---our core contribution---uses a \emph{Geo-guided Appearance Modeling} module that leverages geometric correspondences from the geometry branch to fuse multi-exposure information into a coherent HDR representation.
The outputs of both branches are combined by a Gaussian head to produce HDR 3D Gaussians, which are then converted to LDR via a learned tonemapper (Sec.~\ref{sec:tonemapping}).

\paragraph{\textbf{Geometry Branch.}}
The geometry branch provides the structural foundation for our pipeline.
Following VGGT~\cite{wang2025vggt} and AnySplat~\cite{jiang2025anysplat}, we adopt a pretrained alternating-attention transformer as the geometry backbone.
Each image $I_v$ is patchified into $N = \tfrac{HW}{p^{2}}$ tokens of dimension $d$ using DINOv2~\cite{oquab2023dinov2}, where $p{=}14$ and $d{=}1024$.
To each token sequence $\boldsymbol{t}_v^{G} \in \mathbb{R}^{N \times d}$, we prepend a learnable camera token $\boldsymbol{t}_v^{\text{cam}} \in \mathbb{R}^{1 \times d}$ and four register tokens $\boldsymbol{t}_v^{R} \in \mathbb{R}^{4 \times d}$.
The combined tokens from all $V$ views are processed by an $L$-layer alternating-attention transformer, where each layer applies frame-wise self-attention followed by global cross-view attention.
Dedicated decoder heads for camera poses $p_v$ and depth maps $D_v$ are \emph{frozen} together with the geometry encoder throughout training.

\paragraph{\textbf{Gaussian Head.}}
The Gaussian head combines information from both branches to predict HDR-aware Gaussian attributes.
It takes the geometry tokens $\boldsymbol{t}_v^{G}$ and the high-resolution irradiance features $\boldsymbol{f}_v^{\text{hr}}$ produced by the Geo-guided Appearance Modeling module (Sec.~\ref{sec:appearance}) as input.
The Gaussian head remains \emph{trainable}, enabling it to output HDR-aware Gaussian attributes $\{\sigma_g, \boldsymbol{r}_g, \boldsymbol{s}_g, \boldsymbol{c}_g^{h}\}$.

%% =====================================================
\subsection{Geo-guided Appearance Modeling}
\label{sec:appearance}
The frozen geometry branch provides reliable structure but cannot handle exposure-induced appearance inconsistency.
We introduce the Geo-guided Appearance Modeling module, which mitigates this problem through three stages:
(1)~\emph{Exposure Normalization} aligns inputs to a common reference level,
(2)~\emph{Geo-guided Cross-view Attention} fuses irradiance features using geometric correspondences from the frozen backbone, and
(3)~\emph{High-Resolution Upsampling} recovers pixel-level textures via high-frequency cues.
The output features are decoded into log-radiance SH colors $\boldsymbol{c}_g^{h}$ for each Gaussian.

\paragraph{\textbf{Exposure Normalization $F_E$.}}
To fuse multi-exposure views, we first remove the exposure-induced brightness variation by normalizing all appearance features to a shared reference level.
We define the relative log-exposure of each view as $\tilde{\ell}_v = \ell_v - \bar{\ell}$, where $\bar{\ell}$ is the mid-exposure anchor from Eq.~\eqref{eq:mapping}, and encode $\tilde{\ell}_v$ into a $d$-dimensional embedding $\mathbf{e}_v$ via sinusoidal positional encoding~\cite{mildenhall2021nerf}.
Meanwhile, we extract per-view appearance tokens $\boldsymbol{t}_v^{A} \in \mathbb{R}^{N \times d}$ from each LDR image using a separate patch encoder with the same patch size $p$ and dimension $d$ as the geometry backbone.
A FiLM layer~\cite{perez2018film} then predicts per-view affine parameters:
\begin{equation}
    (\gamma_v,\, \beta_v)
    = \mathrm{FiLM}\!\bigl(\mathbf{e}_v,\;
    \bar{\mathbf{a}}_v,\;
    \bar{\mathbf{a}}\bigr),
\end{equation}
where $\bar{\mathbf{a}}_v = \tfrac{1}{N}\sum_{n=1}^{N} \boldsymbol{t}_{v,n}^{A}$ is the per-view feature mean and $\bar{\mathbf{a}} = \tfrac{1}{V}\sum_{v=1}^{V} \bar{\mathbf{a}}_v$ is the global scene summary.
The appearance tokens are then modulated as:
\begin{equation}
    \hat{\boldsymbol{t}}_v^{A}
    = \boldsymbol{t}_v^{A} \odot (1 + \gamma_v) + \beta_v,
\end{equation}
aligning all views to a shared irradiance level before cross-view fusion.
FiLM is not meant to invert the full non-linear CRF, but to coarsely align exposure distributions into a comparable range before fusion.
The remaining spatially varying effects, which a global FiLM cannot capture, are handled by the subsequent patch-level Geo-guided Attention.
Note that, unlike methods that first linearize inputs via inverse gamma correction~\cite{SelfHDR}, our model operates directly on camera-output LDR images (gamma-encoded), since our training objective reconstructs multi-exposure LDR images rather than linear HDR radiance (Sec.~\ref{sec:loss}).
% Note that unlike methods that first linearize inputs via inverse gamma correction~\cite{SelfHDR}, our model operates directly on camera-output LDR images (which are typically gamma-encoded), since our training objective reconstructs multi-exposure LDR images rather than linear HDR radiance (Sec.~\ref{sec:loss}).

\begin{figure}[t]
\centering
\includegraphics[width=1\linewidth]{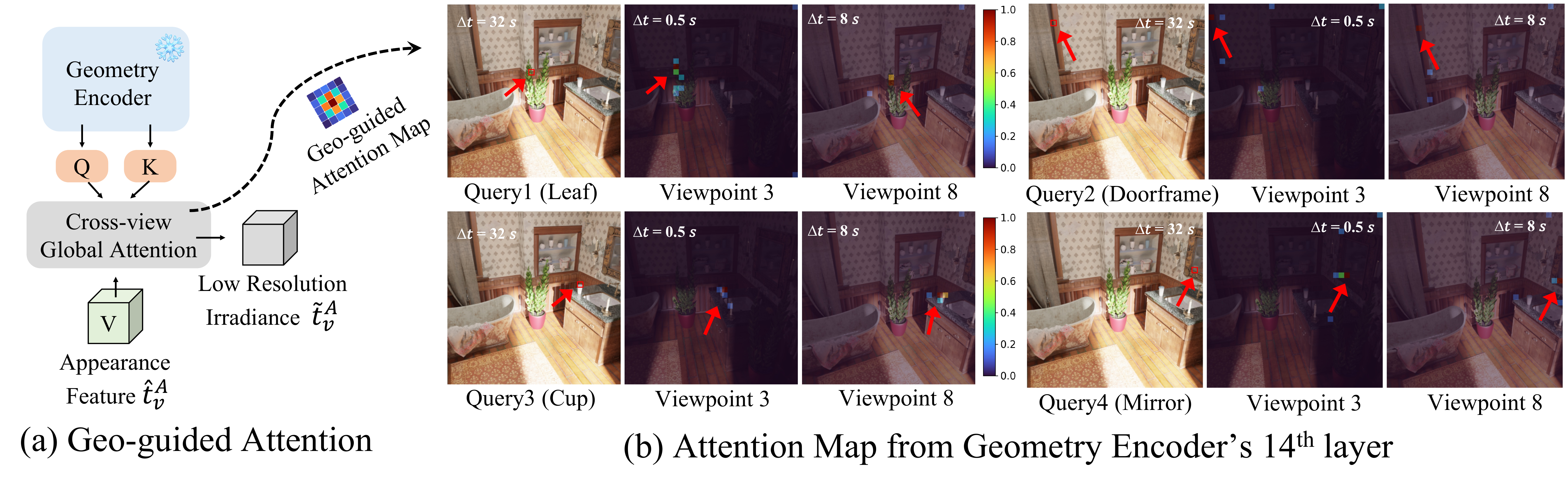}
\caption{
\textbf{Geo-guided Cross-view Attention.} 
(a) The module reuses Q, K from the 14th frozen geometry encoder layer to guide appearance fusion. 
(b) Attention maps visualization shows that it naturally and accurately matches query patches (red box) across views under large viewpoint and extreme exposure variations ($\Delta t$: 0.5--32s).
}
\label{fig:attention_map}
\end{figure}

% \paragraph{\textbf{Geo-guided Cross-view Attention $F_A$.}}
% Different exposures capture complementary information: bright exposures reveal shadow details while dark exposures preserve highlights.
% Fusing this complementary information across views requires cross-view correspondences, which are challenging under large viewpoint and appearance changes and are typically solved by optimization-based feature matching. 
% Interestingly, we observe that \textit{the global attention maps in the frozen geometry encoder already encode reliable cross-view geometric correspondences}, and therefore reuse them to guide appearance fusion (see Fig.~\ref{fig:attention_map}).
% Diverse elements such as leaves, cups, doorframes, and mirrors are accurately matched across views despite extreme exposure variations, greatly benefiting our appearance modeling.
% Concretely, we perform a \emph{cross-attention} operation in which the queries $Q$ and keys $K$ are extracted from the 14th global attention layer of the geometry encoder, while the values are the exposure-normalized appearance tokens $\hat{\boldsymbol{t}}_v^{A}$:
% \begin{equation}
%     \tilde{\boldsymbol{t}}_v^{A}
%     = \mathrm{softmax}\!\left(\frac{Q K^{\!\top}}{\sqrt{d}}\right) \hat{\boldsymbol{t}}_v^{A}.
% \end{equation}

\paragraph{\textbf{Geo-guided Cross-view Attention $F_A$.}}
Different exposures capture complementary information: bright exposures reveal shadows while dark ones preserve highlights.
Fusing them requires cross-view correspondences, which are challenging under large viewpoint and exposure changes.
Interestingly, we observe that \textit{the global attention maps in the frozen geometry encoder already encode reliable cross-view geometric correspondences, greatly benefiting our appearance modeling}.
Therefore, we reuse these attention maps to guide appearance fusion.
As shown in Fig.~\ref{fig:attention_map} (b), diverse elements such as leaves, cups, doorframes, and mirrors are accurately matched across views despite extreme exposure variations.
\begin{equation}
    \tilde{\boldsymbol{t}}_v^{A}
    = \mathrm{softmax}\!\left(\frac{Q K^{\!\top}}{\sqrt{d}}\right) \hat{\boldsymbol{t}}_v^{A}.
\end{equation}

\paragraph{\textbf{High-Resolution Upsampling $F_U$.}}
The irradiance features $\tilde{\boldsymbol{t}}_v^{A}$ operate at patch resolution ($\tfrac{H}{p} \times \tfrac{W}{p}$), losing high-frequency details critical for realistic appearance.
To recover pixel-level textures while preserving the fused low-frequency irradiance, we adopt a Difference-of-Gaussians (DoG) guided upsampling strategy.
We first encode each full-resolution LDR image into a feature map $\mathbf{g}_v \in \mathbb{R}^{d'\times H \times W}$ via a shallow CNN.
A low-passed version $\mathbf{g}_v^{\downarrow\uparrow}$ is obtained by downsampling and bilinearly upsampling $\mathbf{g}_v$, yielding the high-frequency residual $\mathbf{g}_v - \mathbf{g}_v^{\downarrow\uparrow}$.
This residual is then added to the bilinearly upsampled irradiance features to produce pixel-level irradiance features:
\begin{equation}
    \boldsymbol{f}_v^{\text{hr}}
    = \mathrm{Conv}\!\bigl(
      \mathrm{Up}(\tilde{\boldsymbol{t}}_v^{A})
      + \mathrm{Conv}(\mathbf{g}_v - \mathbf{g}_v^{\downarrow\uparrow})
    \bigr)
    \;\in\; \mathbb{R}^{d \times H \times W},
\label{eq:upsample}
\end{equation}
combining multi-view irradiance consensus with per-image structural detail.

\paragraph{\textbf{HDR 3D Gaussian Prediction.}}
We now merge the geometry and appearance branches to produce HDR 3D Gaussians.
A DPT decoder~\cite{ranftl2021vision} upsamples the geometry tokens $\boldsymbol{t}_v^{G}$ to pixel resolution, and the result is added to the irradiance features $\boldsymbol{f}_v^{\text{hr}}$.
A lightweight CNN then regresses per-Gaussian opacity, orientation, scale, and log-radiance SH color:
\begin{equation}
    \{\sigma_g,\, \boldsymbol{r}_g,\, \boldsymbol{s}_g,\, \boldsymbol{c}_g^{h}\}
    = F_G\!\bigl(\mathrm{DPT}(\boldsymbol{t}_v^{G}) + \boldsymbol{f}_v^{\text{hr}}\bigr).
\end{equation}
The Gaussian centers $\boldsymbol{\mu}_g$ are obtained by back-projecting the predicted depth maps $D_v$ through the estimated camera poses $p_v$.
The Gaussians are then voxelized~\cite{jiang2025anysplat} to reduce primitive count for efficient splatting.

%% =====================================================
\subsection{3D HDR-to-2D LDR Mapping}
\label{sec:tonemapping}

Given HDR 3D Gaussians, rendering a photorealistic LDR view requires recovering the Camera Response Function (CRF) that maps scene irradiance to observed pixel values.
Unlike optimization-based methods that overfit a per-scene MLP tonemapper~\cite{liu2025gausshdr,cai2024hdr}, our method seeks a \emph{generalizable} tonemapper that adapts to different cameras without per-scene optimization.
We achieve this goal via a \textbf{Meta Net} $F_M$ that predicts the parameters of a lightweight tonemapper from scene context.

\paragraph{\textbf{Tone Mapping Formulation.}}
We convert HDR radiance to LDR in two steps.
First, Gaussian splatting rasterizes the linear radiance into a per-pixel HDR image at view $v$:
\begin{equation}
    \mathbf{H}_v
    = \mathcal{R}\!\bigl(\exp(\boldsymbol{c}_g^{h}),\,
      \sigma_g,\, \boldsymbol{r}_g,\, \boldsymbol{s}_g,\, \boldsymbol{\mu}_g;\; p_v\bigr)
    \;\in\; \mathbb{R}^{H \times W \times 3},
\label{eq:rasterize}
\end{equation}
where $\mathcal{R}(\cdot)$ denotes the differentiable rasterization with alpha-weighted blending~\cite{3dgs} in linear radiance space, $p_v$ is the camera pose, and the $\exp$ converts log-radiance SH colors $\boldsymbol{c}_g^{h}$ back to linear radiance before blending.

Second, following the log-domain CRF model~\cite{malik}, a learned tonemapper $g_{\boldsymbol{\theta}}$ maps the log-irradiance to $[0,1]$ LDR values:
\begin{equation}
    \mathbf{L}_v(\ell)
    = g_{\boldsymbol{\theta}}\!\bigl(
      \log \mathbf{H}_v
      + (\ell - \bar{\ell})\cdot\log 2
    \bigr),
\label{eq:tonemap}
\end{equation}
where $\mathbf{L}_v(\ell) \in \mathbb{R}^{H \times W \times 3}$ is the rendered LDR image at view $v$ under target log-exposure $\ell$, $\bar{\ell}$ is the mid-exposure anchor from Eq.~\eqref{eq:mapping}, and the term $(\ell - \bar{\ell})\cdot\log 2$ adjusts the irradiance to the desired exposure level.
Here $g_{\boldsymbol{\theta}}$ is a two-layer MLP with hidden dimension $h$ (input: $3 \to$ hidden: $h$ with ReLU $\to$ output: $3$ with sigmoid), acting as a learnable inverse CRF.
Rather than learning a fixed $\boldsymbol{\theta}$, we predict \emph{scene-specific} parameters via the Meta Net, enabling adaptation to different cameras and tone curves without per-scene optimization.

\paragraph{\textbf{Meta Net $F_M$.}}
The Meta Net infers scene-specific tonemapper parameters $\boldsymbol{\theta}$ in a single forward pass, enabling $g_{\boldsymbol{\theta}}$ to reproduce the original camera's tone curve without per-scene optimization.
% For brevity, we denote the full set of pixel-level Gaussians (before voxelization) as $\mathcal{G} = \{(\boldsymbol{\mu}_g, \sigma_g, \boldsymbol{r}_g, \boldsymbol{s}_g, \boldsymbol{c}_g^{h})\}_{g=1}^{G}$, where $G = V \!\times\! H \!\times\! W$.
For brevity, we denote the full set of pixel-level Gaussians (before voxelization) as:
\begin{equation}
    \mathcal{G} = \bigl\{(\boldsymbol{\mu}_g,\, \sigma_g,\, \boldsymbol{r}_g,\, \boldsymbol{s}_g,\, \boldsymbol{c}_g^{h})\bigr\}_{g=1}^{G},
    \quad G = V \!\times\! H \!\times\! W.
\end{equation}
The Meta Net takes three inputs:
(i)~the full-resolution LDR features $\mathbf{g}_v$ from the upsampling CNN (Eq.~\eqref{eq:upsample}),
(ii)~the per-view exposure embeddings $\{\mathbf{e}_v\}_{v=1}^{V}$, and
(iii)~the predicted HDR Gaussians $\mathcal{G}$.
These are concatenated and compressed by a strided convolutional encoder and then globally pooled across all spatial and views dimensions to produce a scene-level descriptor $\boldsymbol{\theta}$:
\begin{equation}
    \boldsymbol{\theta}
    = F_M\!\bigl(\{\mathbf{g}_v\},\, \{\mathbf{e}_v\},\, \mathcal{G}\bigr)
    \;\in\; \mathbb{R}^{d_{\theta}},
\end{equation}
where $d_{\theta}$ encodes all weights and biases of the two-layer tonemapper $g_{\boldsymbol{\theta}}$.

%% =====================================================
\subsection{Training Strategies}
\label{sec:loss}

\paragraph{\textbf{Training Objective.}}
InstantHDR is trained end-to-end without any 3D or HDR supervision, using only multi-view LDR images with known exposure times.
The geometry encoder and its decoder heads remain frozen; only the appearance branch, the Gaussian head, and the Meta Net are optimized.

During training, the target views are the same with the context views, i.e., the model predicts HDR Gaussians $\mathcal{G}$ and tonemapper parameters $\boldsymbol{\theta}$ from $\{I_v, \ell_v\}_{v=1}^{V}$ and is supervised by rendering back to the same views and exposures via Eqs.~\eqref{eq:rasterize}--\eqref{eq:tonemap}.
At test time, the target view and exposure can differ from the context set, enabling novel-view synthesis at arbitrary exposures.
The total loss is:
\begin{equation}
    \mathcal{L} = \mathcal{L}_{\text{RGB}} + \lambda_g \,\mathcal{L}_g.
\end{equation}
The photometric loss compares rendered and ground-truth LDR images:
\begin{equation}
    \mathcal{L}_{\text{RGB}}
    = \frac{1}{V}\sum_{v=1}^{V}
    \Bigl[
      \mathrm{MSE}\bigl(I_v,\, \mathbf{L}_v(\ell_v)\bigr)
      + \lambda_{\text{perc}} \cdot \mathcal{L}_{\text{perc}}\bigl(I_v,\, \mathbf{L}_v(\ell_v)\bigr)
    \Bigr].
\end{equation}
The geometry consistency loss enforces alignment between the depth maps $D_v$ from the frozen DPT head and the rendered depth maps $\hat{D}_v$ from the predicted Gaussians.
Since $D_v$ can be unreliable in challenging regions (\eg, sky or reflective surfaces), we utilize the jointly learned confidence map $C_v^{D}$ and apply supervision only to the top-$N\%$ most confident pixels:
\begin{equation}
    \mathcal{L}_g = \frac{1}{V}\sum_{v=1}^{V} \bigl(D_v[M_v] - \hat{D}_v[M_v]\bigr)^2,
\end{equation}
where $M_v$ is a binary mask corresponding to the top-$N$ quantile of $C_v^{D}$; we set $N{=}30$ in all experiments.
Our model learns HDR representations implicitly by correctly reconstructing LDR images across different exposure times.

\paragraph{\textbf{Post-Optimization.}}
We can refine the predicted Gaussians and camera parameters via lightweight post-optimization.
After pruning low-opacity Gaussians ($\sigma < 0.01$), we minimize a combination of MSE and SSIM losses between rendered and input images for 1K iterations, back-propagating gradients through all Gaussian attributes, and tonemapper parameters.
The learning rates are: 1.6e-4 (position), 5e-3 (scale), 1e-3 (rotation), 5e-2 (opacity), and 2.5e-3 (color).

% \paragraph{\textbf{Evaluation-Time Pose Alignment.}}
% Since InstantHDR predicts poses rather than assuming known ones, the reconstructed scene may differ from the ground truth by a rigid transformation.
% To fairly compare with pose-dependent baselines~\cite{liu2025gausshdr,cai2024hdr}, we follow prior pose-free works~\cite{ye2024no} and align poses at test time: we freeze $\mathcal{G}$ and $\boldsymbol{\theta}$, then optimize only the target camera pose so that $\mathbf{L}_v(\ell_v)$ matches $I_v$.
% This step is solely for evaluation and is not needed in practice.

\section{Experiments}

% 1. Pretraining Datasets / Dataset Vis / Table
% Implementation Details.
% Evaluation Metrics.

% 2-1. Main LDR Compare / LDR Vis
% 2-2. 
% 3. HDR Compare / HDR Vis
% 4. Ablation - Exp Norm / Cross-view Attn / Upsampling

\begin{table}[t]
\centering
\caption{Comparison of existing HDR datasets for novel view synthesis. ``Pano.'' denotes panoramic 360$^\circ$ images. ``Multi-Exp.'' indicates multi-exposure LDR images, with the number of exposures in parentheses.}
\resizebox{1\linewidth}{!}{
\renewcommand{\arraystretch}{0.6}
\begin{tabular}{lcccccc}
\toprule
Dataset & Source & \#Scenes & Type & HDR GT & Multi-Exp. & Depth/Normal \\
\midrule
HDR-NeRF~\cite{huang2022hdr} & Blender + Real & 8 + 4 & Object/Indoor & \checkmark & \checkmark~(5) & \texttimes \\
HDR-Plenoxels~\cite{jun2022hdr} & Blender + Real & 5 + 4 & Indoor & \texttimes & \checkmark~(3) & \texttimes \\
Pano-NeRF~\cite{lu2024pano} & Blender + Replica + Real & 5 + 8 + 3 & Indoor (Pano.) & \checkmark & \checkmark~(9) & \checkmark \\
PanDORA~\cite{dastjerdi2024pandora} & Real capture (360$^\circ$) & 14 & Indoor (Pano.) & \checkmark & \checkmark~(2) & \texttimes \\
\midrule
HDR-Pretrain (Ours) & Blender & \textbf{168} & Indoor & \checkmark & \checkmark~(5) & \checkmark \\
\bottomrule
\end{tabular}
\label{tab:hdr_datasets}
}
\end{table}

\begin{figure}[t]
\centering
\includegraphics[width=\linewidth]{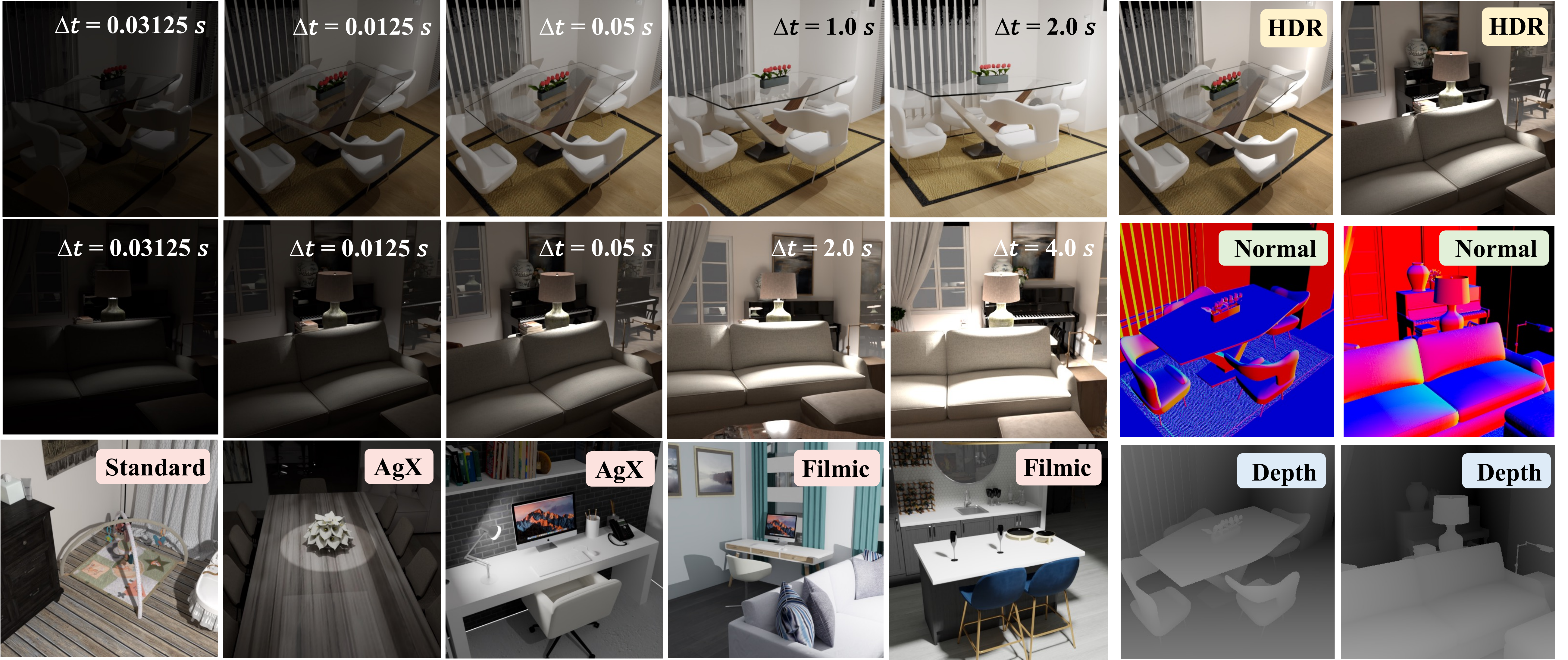}
\caption{
\textbf{Examples from our HDR-Pretrain dataset.} 
Each scene includes multi-view, multi-exposure LDR images at varying $\Delta t$, 32-bit HDR ground truth, depth and normal maps, rendered under diverse tone-mapping operators (Standard, AgX, Filmic).
}
\label{fig:HDR_Pretrain}
\end{figure}

\noindent\textbf{Pretraining Dataset.}
HDR datasets with multi-view LDR images remain extremely scarce.
As shown in Tab.~\ref{tab:hdr_datasets}, existing benchmarks contain only a handful of scenes, far from sufficient for large-scale pretraining.
To bridge this gap, we construct \textbf{HDR-Pretrain}, in Fig.~\ref{fig:HDR_Pretrain}, a large-scale synthetic dataset of 168 photorealistic indoor scenes rendered in Blender.
The 3D assets are sourced from HSSD~\cite{khanna2024habitat}, an open-source collection of realistic interiors originally built for embodied AI research.
Following HDR-NeRF~\cite{huang2022hdr}, we sample viewpoints on a $5\times7$ grid with $2.5^\circ$ / $5^\circ$ angular steps per scene and render 32-bit HDR images via Cycles path tracing at $448\times448$ resolution.
Each view is paired with 5 exposure-bracketed LDR images under a randomly chosen tone-mapping operator, as well as depth and normal maps.
We randomly apply one of three tone-mapping operators (AgX, Filmic, Standard) per scene to increase data diversity.

\noindent\textbf{Evaluation Dataset.}
We mainly evaluate on the HDR-NeRF benchmark~\cite{huang2022hdr}, which contains 8 synthetic and 4 real indoor scenes, each captured from 35 viewpoints at 5 exposure levels $\{t_1, t_2, t_3, t_4, t_5\}$.
Following the standard protocol~\cite{huang2022hdr}, 18 views with one exposure drawn from $\{t_1, t_3, t_5\}$ are used as input, while the remaining views are held out for evaluation.
For clarity, we report the average LDR metrics across all 5 exposures rather than separating observed (LDR-OE) and novel (LDR-NE) exposures.
We further test under sparse-view settings with only 4 or 8 input views.

% \noindent\textbf{Implementation Details.}
% We build upon AnySplat~\cite{jiang2025anysplat}, initializing from its pretrained weights with the most backbone frozen during training.
% The voxel size for differentiable voxelization is set to $\epsilon=0.002$.
% We train the model using AdamW with a cosine learning rate scheduler, a peak learning rate of $2\times10^{-4}$, and a warmup of 1K iterations.
% Training runs for $30K$ iterations on a single node with bf16 mixed precision.
% The input resolution is $448\times448$, and we apply intrinsic augmentation via random center-cropping and horizontal flipping.
% For the training objective, we set $\lambda_{\text{perc}}=0.05$ and $\lambda_{\text{g}}=0.1$.
% During each iteration, we randomly sample $2\sim10$ context views per scene.
% We train InstantHDR on 8 NVIDIA A6000 GPUs for approximately two days.
% Due to the domain gap between real and our synthetic HDR-Pretrain scenes, we finetune on HDR-Plenoxels 4 real scenes before evaluating on HDR-NeRF real scenes, ensuring the testing scenes is always unseen.
% Following other pose-free methods~\cite{ye2024no}, we perform test-time pose alignment for evaluation and post optimization.

\noindent\textbf{Implementation Details.}
We build upon AnySplat~\cite{jiang2025anysplat} with its backbone frozen, using a voxel size of $\epsilon{=}0.002$ in normalized scene coordinates.
Training uses AdamW with cosine scheduling, peak learning rate $2{\times}10^{-4}$, 1K warmup, and runs for 30K iterations in bf16 precision on 8 NVIDIA A6000 GPUs for $\sim$2 days.
Input resolution is $448{\times}448$ with random cropping and flipping augmentation.
We set $\lambda_{\text{perc}}{=}0.05$ and $\lambda_{\text{g}}{=}0.1$, sampling $2{\sim}10$ context views per iteration.
Due to the domain gap between real and our synthetic HDR-Pretrain scenes, we finetune on HDR-Plenoxels 4 real scenes before evaluating on HDR-NeRF real scenes, ensuring all testing scenes are unseen.
Following pose-free methods~\cite{ye2024no}, we perform test-time pose alignment for evaluation and post-optimization.

\noindent\textbf{Evaluation Metrics.}
All images are resized to $448\times 448$ for fair comparison.
We report PSNR, SSIM, LPIPS~\cite{zhang2018unreasonable}, and reconstruction time for quantitative and efficiency comparison.
Following HDR-NeRF~\cite{huang2022hdr}, HDR results are evaluated in the tone-mapped domain using the $\mu$-law~\cite{hdr_26}, with 99th-percentile normalization to suppress extreme HDR values.

\subsection{Quantitative Results}

\noindent\textbf{LDR Comparisons on Zero-shot Inference.}
% We first evaluate zero-shot cross-domain performance on HDR-NeRF~\cite{huang2022hdr} scenes.
% Existing feed-forward models such as AnySplat~\cite{jiang2025anysplat} do not account for exposure inconsistency across input views, leading to severely degraded reconstruction under varying exposures.
% As shown in Tab.~\ref{tab:hdr_nerf}, our InstantHDR consistently outperforms AnySplat by a large margin across all settings.
% On the real scenes with 8 input views, InstantHDR surpasses AnySplat by +5.65 dB in PSNR, +0.155 in SSIM, and $-$0.167 in LPIPS, while maintaining comparable reconstruction time within 1.582 seconds.
% The gap is even more pronounced on synthetic scenes, where InstantHDR exceeds AnySplat by +8.07 dB in PSNR and +0.256 in SSIM with 8 views.
% These results demonstrate that InstantHDR generalizes well to unseen scenes with exposure-inconsistent inputs while maintaining fast reconstruction speed.
We first evaluate zero-shot results on HDR-NeRF~\cite{huang2022hdr} scenes.
AnySplat~\cite{jiang2025anysplat} ignores exposure inconsistency, leading to severely degraded results.
As shown in Tab.~\ref{tab:hdr_nerf}, InstantHDR consistently outperforms AnySplat by large margins (\textit{e.g.}, +5.65 dB on real scenes, +8.07 dB on synthetic scenes with 8 views) with negligible increase in reconstruction time, showing strong generalization to exposure-inconsistent inputs.

\noindent\textbf{LDR Comparisons on Optimization.}
We further compare against state-of-the-art optimization-based methods, HDR-GS~\cite{cai2024hdr} and GaussHDR~\cite{liu2025gausshdr}.
In this setting, we post-optimize the Gaussians and tone-mapping parameters produced by InstantHDR and AnySplat for 1K iterations, denoted as InstantHDR\_1K and AnySplat\_1K. 
Under the challenging 4-view sparse setting on real scenes, InstantHDR\_1K achieves 22.16 dB PSNR and 0.762 SSIM, surpassing GaussHDR by +2.90 dB and +0.071 in SSIM, demonstrating that the diverse geometric priors from feed-forward foundation models greatly benefit sparse-view reconstruction.
Under the denser 18-view setting, our method remains competitive on real scenes while showing a modest gap on synthetic scenes.
Notably, InstantHDR\_1K requires only $\sim$30--40 seconds per scene, roughly \textbf{20$\times$} faster than HDR-GS and \textbf{50$\times$} faster than GaussHDR, attributing to the good feed-forward initialization and the ability to skip the costly iterative densification.

\begin{table}[t]
\centering
\caption{Quantitative LDR comparison on HDR-NeRF~\cite{huang2022hdr} real and synthetic scenes with varying numbers of input views. 
We compare our method, as well as its variants with 1K iters of post-optimization (Ours\_1K), against AnySplat~\cite{jiang2025anysplat} HDR-GS~\cite{cai2024hdr} and GaussHDR~\cite{liu2025gausshdr}.
Time denotes per-scene reconstruction time in seconds.
}
\label{tab:hdr_nerf}
\resizebox{\textwidth}{!}{
\renewcommand{\arraystretch}{0.8}
\begin{tabular}{cl cccc cccc cccc}
\toprule[2pt]
\multirow{2}{*}{Mode} & \multirow{2}{*}{Method} & \multicolumn{4}{c}{4 Views} & \multicolumn{4}{c}{8 Views} & \multicolumn{4}{c}{18 Views} \\
\cmidrule(lr){3-6} \cmidrule(lr){7-10} \cmidrule(lr){11-14}
 &  & PSNR$\uparrow$ & SSIM$\uparrow$ & LPIPS$\downarrow$ & Time(s)$\downarrow$ & PSNR$\uparrow$ & SSIM$\uparrow$ & LPIPS$\downarrow$ & Time(s)$\downarrow$ & PSNR$\uparrow$ & SSIM$\uparrow$ & LPIPS$\downarrow$ & Time(s)$\downarrow$ \\
\midrule
\multicolumn{14}{l}{\textit{HDR-NeRF Real Dataset}~\cite{huang2022hdr}} \\
\multirow{2}{*}{Zero-shot}
& AnySplat~\cite{jiang2025anysplat} & 12.10 & 0.517 & 0.497 & \textbf{0.973} & 13.30 & 0.569 & 0.436 & \textbf{1.180} & 13.91 & 0.600 & 0.403  & \textbf{2.139} \\
& InstantHDR (Ours) & \textbf{18.44} & \textbf{0.721} & \textbf{0.269} & 1.033 & \textbf{18.95} & \textbf{0.724} & \textbf{0.269} & 1.582 & \textbf{19.48} & \textbf{0.745} & \textbf{0.257} & 2.512 \\
\midrule
\multirow{4}{*}{Optimization}
& HDR-GS~\cite{cai2024hdr} & 15.40 & 0.622 & 0.334 & 872 & 23.02 & 0.791 & 0.121 & 736 & 27.42 & 0.893 & 0.047 & 815 \\
& GaussHDR~\cite{liu2025gausshdr} & 19.26 & 0.691 & 0.270 & 1833 & 24.96 & \textbf{0.854} & \textbf{0.068} & 1816 & \textbf{29.36} & 0.929 & \textbf{0.024} & 1891 \\
& AnySplat\_1K~\cite{jiang2025anysplat} & 11.84 & 0.486 & 0.468 & \textbf{30} & 13.63 & 0.580 & 0.372 & \textbf{33} & 14.86 & 0.689 & 0.266 & 40 \\
& InstantHDR\_1K (Ours) & \textbf{22.16} & \textbf{0.762} & \textbf{0.259} & 32 & \textbf{25.32} & 0.852 & 0.160 & 40 & 29.19 & \textbf{0.931} & 0.086 & \textbf{39} \\
\midrule[2pt]
\multicolumn{14}{l}{\textit{HDR-NeRF Syn Dataset}~\cite{huang2022hdr}} \\
\multirow{2}{*}{Zero-shot}
& AnySplat~\cite{jiang2025anysplat} & 13.98 & 0.525 & 0.400 & \multirow{2}{*}{-}& 14.51 & 0.573 & 0.378 & \multirow{2}{*}{-} & 15.27 & 0.534 & 0.327 & \multirow{2}{*}{-} \\
& InstantHDR (Ours) & \textbf{21.76} & \textbf{0.728} & \textbf{0.172} &  & \textbf{22.58} & \textbf{0.785} & \textbf{0.138} &  & \textbf{22.59} & \textbf{0.830} & \textbf{0.115} &  \\
\midrule
\multirow{4}{*}{Optimization}
& HDR-GS~\cite{cai2024hdr} & 24.26 & 0.711 & 0.210 & \multirow{4}{*}{-} & 30.60 & 0.867 & 0.086 & \multirow{4}{*}{-} & 29.93 & 0.917 & 0.061 & \multirow{4}{*}{-} \\
& GaussHDR~\cite{liu2025gausshdr} & 21.62 & 0.646 & 0.224 &  & \textbf{34.49} & \textbf{0.924} & \textbf{0.026} &  & \textbf{38.63} & \textbf{0.969} & \textbf{0.009} &  \\
& AnySplat\_1K~\cite{jiang2025anysplat} & 14.01 & 0.526 & 0.347 &  & 15.42 & 0.656 & 0.248 &  & 16.39 & 0.624 & 0.221 &  \\
& InstantHDR\_1K (Ours) & \textbf{27.63} & \textbf{0.825} & \textbf{0.137} &  & 32.75 & 0.922 & 0.061 &  & 35.99 & 0.965 & 0.037 &  \\
\bottomrule[2pt]
\end{tabular}
}
\end{table}

\begin{table}[t]
\centering
\caption{
(a) Quantitative HDR comparison on HDR-NeRF~\cite{huang2022hdr} synthetic scenes with 8 input views, evaluated in the $\mu$-law tone-mapped domain.
(b) Ablation study of InstantHDR. Zero-shot results on HDR-NeRF~\cite{huang2022hdr} real scenes with 8 input views.
}
% \label{tab:hdr_plenoxels}
% \setlength{\tabcolsep}{4pt}
\resizebox{\textwidth}{!}{
\begin{tabular}{cc}

\renewcommand{\arraystretch}{0.9}
\begin{tabular}{cl cccc}
\multicolumn{6}{c}{\fontsize{11}{11}\selectfont(a) HDR Comparison on HDR-NeRF~\cite{huang2022hdr} syn scenes.} \\
\toprule[2pt]
Mode & Method & PSNR$\uparrow$ & SSIM$\uparrow$ & LPIPS$\downarrow$ & Time(s)$\downarrow$ \\
\midrule
\multirow{2}{*}{Zero-shot}
& AnySplat & 8.93 & 0.595 & 0.416 & \textbf{1.180}\\
& InstantHDR (Ours) & \textbf{15.29} & \textbf{0.772} & \textbf{0.140} & 1.582 \\
\midrule
\multirow{4}{*}{Optimization}
& HDR-GS & 27.69 & 0.871 & 0.090 & 736 \\
& GaussHDR & \textbf{31.62} & 0.887 & \textbf{0.037} & 1816 \\
& AnySplat\_1K & 9.52 & 0.678 & 0.268 & \textbf{33} \\
& InstantHDR\_1K (Ours) & 27.55 & \textbf{0.899} & 0.076 & 40 \\
\bottomrule[2pt]
\end{tabular}

& 

\renewcommand{\arraystretch}{1.1}
\begin{tabular}{l ccc}
\multicolumn{4}{c}{\fontsize{11}{11}\selectfont(b) Ablation on HDR-NeRF~\cite{huang2022hdr} real scenes.} \\
\toprule[2pt]
Method & PSNR$\uparrow$ & SSIM$\uparrow$ & LPIPS$\downarrow$ \\
\midrule
w/o Meta Net & 16.32 & 0.699 & 0.289 \\
w/o Exposure Norm & 13.72 & 0.693 & 0.278 \\
w/o Cross-view Attn & 17.63 & 0.702 & 0.277 \\
w/o Upsampling & \textbf{19.20} & 0.718 & 0.386 \\
Ours & 18.95 & \textbf{0.724} & \textbf{0.269} \\
\bottomrule[2pt]
\end{tabular}

\label{tab:hdr_syn}
\end{tabular}
}
\end{table}

% \begin{figure}[t]
% \centering
% % \includegraphics[width=\linewidth]{asset/LDR-syn.pdf}
% \includegraphics[height=6cm]{asset/LDR-syn.pdf}
% \caption{
% \textbf{LDR visual comparisons on the synthetic scenes.} 
% Feed-forward methods~\cite{jiang2025anysplat} fail on multi-exposure inputs, while optimization-based methods~\cite{liu2025gausshdr} require $\sim$2K seconds per scene. Our InstantHDR achieves competitive quality in under 40s. Yellow/blue tags denote reconstruction time/PSNR.
% }
% \label{fig:ldr-syn}
% \end{figure}

% \begin{figure}[t]
% \centering
% \includegraphics[height=6cm]{asset/LDR-real.pdf}
% \caption{
% \textbf{LDR visual comparisons on real scenes.} Our InstantHDR generalizes well to real-world captures.
% Blue tags denote reconstruction PSNR.\
% }
% \label{fig:ldr-real}
% \end{figure}

\noindent\textbf{HDR Comparisons.}
Tab.~\ref{tab:hdr_syn} (a) reports HDR results on HDR-NeRF synthetic scenes.
Although never supervised with HDR ground truth, InstantHDR implicitly recovers HDR from multi-exposure LDR inputs, outperforming AnySplat by +6.36 dB in PSNR in zero-shot mode.
With 1K steps of post-optimization, InstantHDR\_1K reaches 27.55 dB PSNR and 0.899 SSIM, achieving comparable performance to HDR-GS (27.69 dB) while surpassing GaussHDR in SSIM.
The remaining gap with GaussHDR in PSNR is likely due to its dedicated 3D--2D dual-branch tone-mapping design, whereas both our method and HDR-GS adopt a simpler single tone mapping branch.
We believe that incorporating more advanced tone-mapping modules is a promising avenue for future works.

\subsection{Qualitative Results}

\begin{figure}[H]
\centering
\begin{subfigure}[t]{\linewidth}
  \centering
  \includegraphics[width=\linewidth]{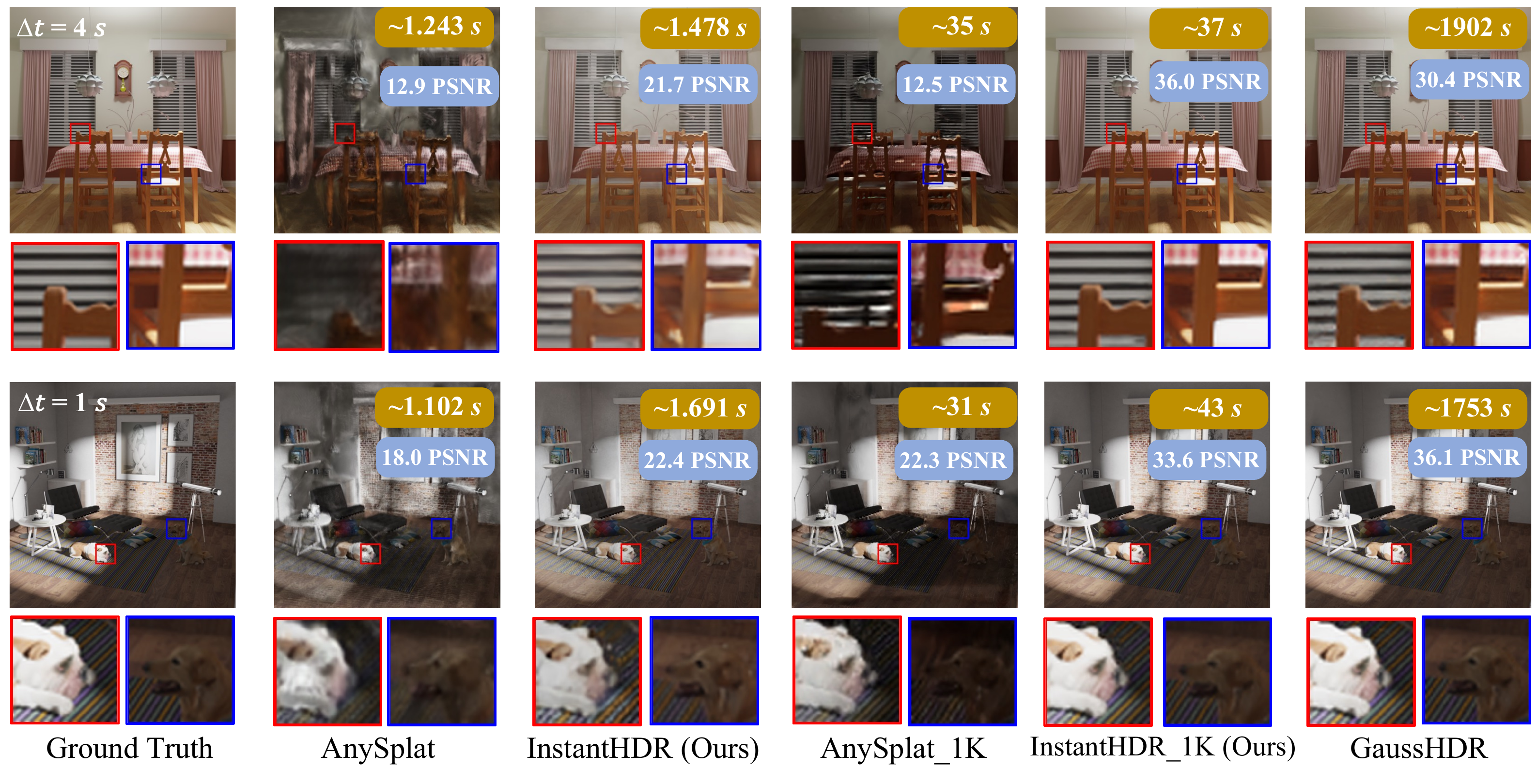}
  \caption{LDR visual comparisons on HDR-NeRF~\cite{huang2022hdr} synthetic scenes.}
  \label{fig:ldr-syn}
\end{subfigure}\\[4pt]
\begin{subfigure}[t]{\linewidth}
  \centering
  \includegraphics[width=\linewidth]{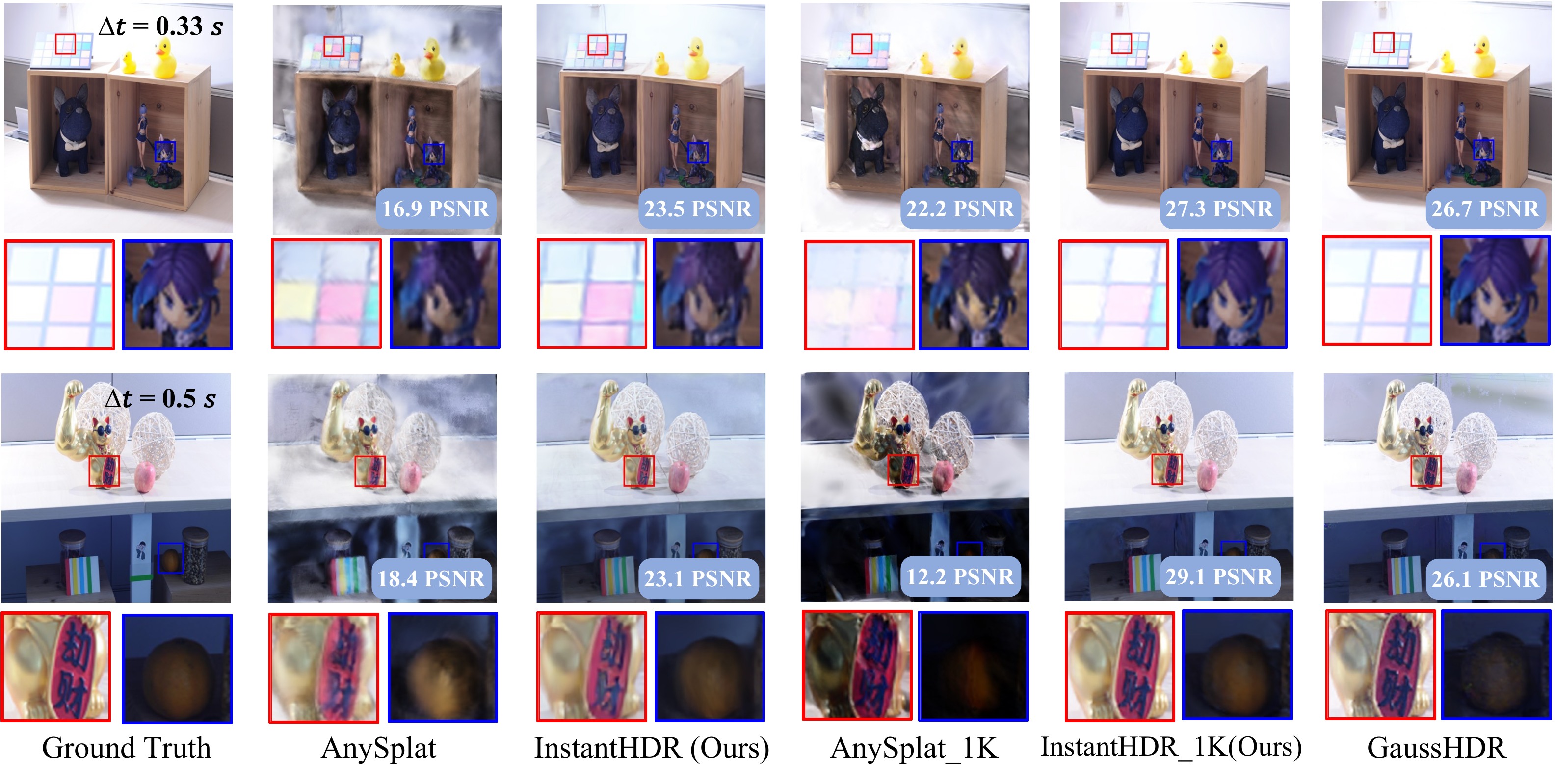}
  \caption{LDR visual comparisons on HDR-NeRF~\cite{huang2022hdr} real scenes.}
  \label{fig:ldr-real}
\end{subfigure}
\caption{\textbf{LDR visual comparisons.}
Feed-forward methods~\cite{jiang2025anysplat} fail on multi-exposure inputs, while optimization-based methods~\cite{liu2025gausshdr} require $\sim$2K seconds per scene. Our InstantHDR achieves competitive quality in under 40s. Yellow/blue tags denote reconstruction time/PSNR.}
\label{fig:ldr}
\end{figure}

\noindent\textbf{LDR Novel View Rendering.}
In Fig.~\ref{fig:ldr}, AnySplat struggles in exposure variations, while GaussHDR is time-consuming. 
Our InstantHDR generalizes well to both synthetic and real-world scenes, renders clean, exposure-controllable LDR views in seconds, and achieves competitive quality after 1K post-optimization.

\begin{figure}[t]
\centering
\includegraphics[height=4.5cm]{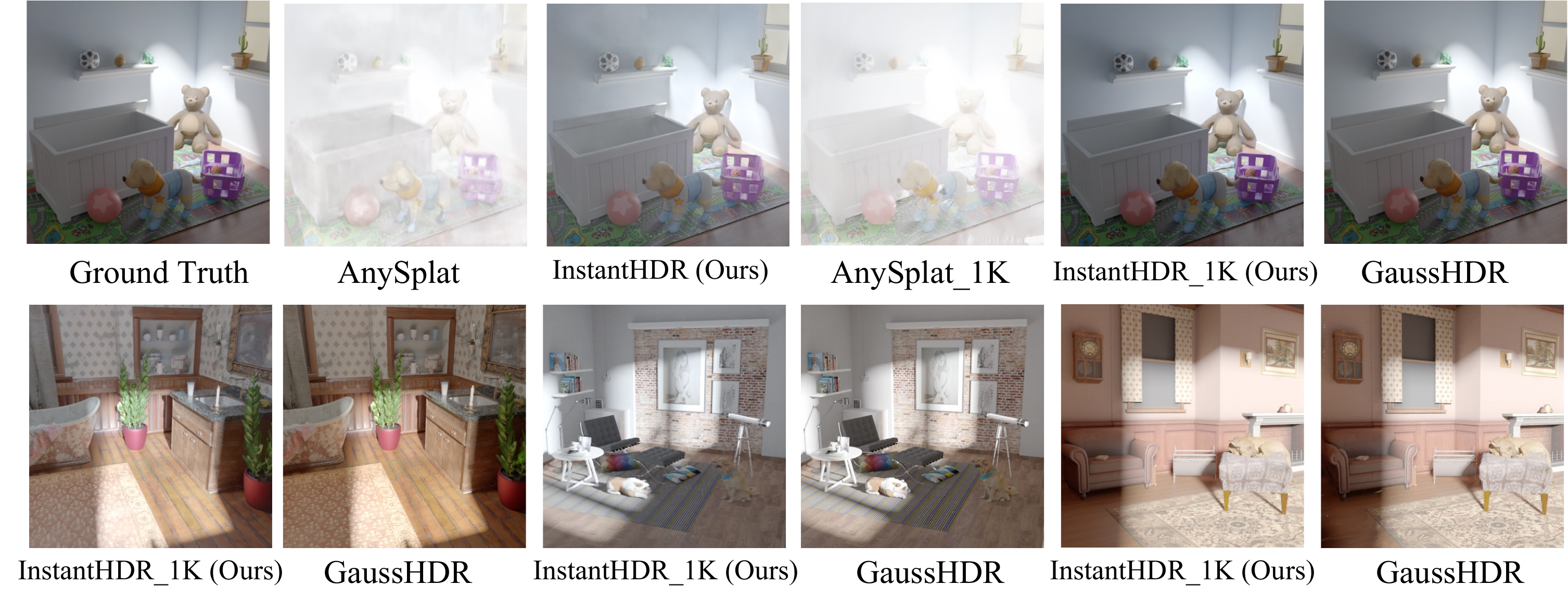}
\caption{
\textbf{HDR visual comparisons.} After 1K post-optimization, our InstantHDR produces HDR results comparable to time-consuming GaussHDR.
}
\label{fig:hdr}
\end{figure}

\noindent\textbf{HDR Novel View Rendering.}
As shown in Fig.~\ref{fig:hdr}, zero-shot HDR outputs from feed-forward models (AnySplat \& InstantHDR) appear overly bright, as extreme radiance values are hard to predict accurately in a single-forward pass, which inflates the average brightness after normalization.
We see this as an open challenge for feed-forward HDR models. 
After 1K post-optimization, this issue is largely alleviated, with our InstantHDR producing results similar to GaussHDR.

\begin{figure}[t]
\centering
\includegraphics[height=2.5cm]{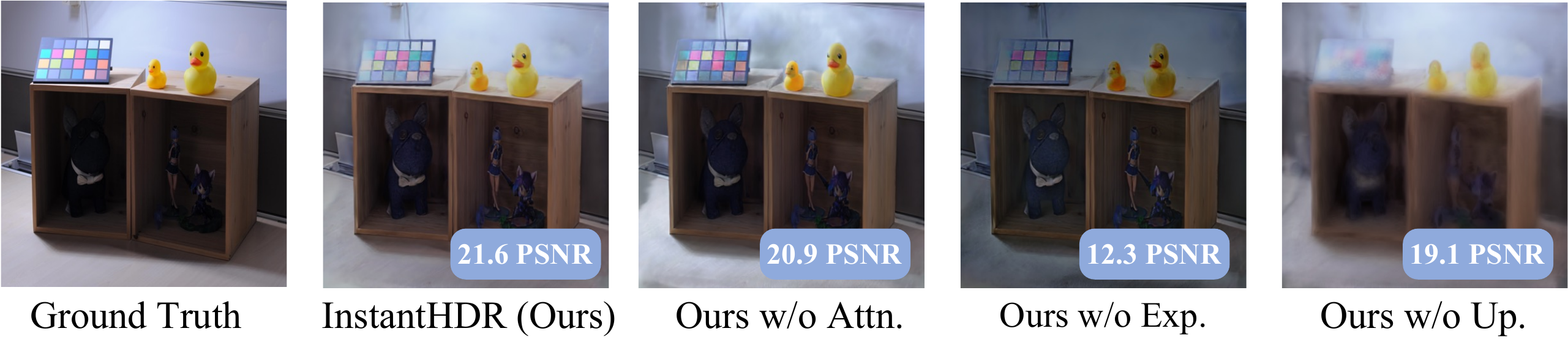}
% \vskip-10pt\caption{
\caption{
\textbf{Ablation visualization.} Attn.=cross-view attention, Exp.=exposure normalization, Up.=upsampling. Removing Attn.\ causes wall ghosting; removing Exp.\ shifts overall brightness; removing Up.\ produces blurry results.
}
\label{fig:ablation}
\end{figure}

% \subsection{Ablation Study}
% %As shown in Tab.~\ref{tab:hdr_syn}~(b) and Fig.~\ref{fig:ablation}, each component contributes meaningfully.
% %Without Meta Net, training becomes harder to converge as the model cannot adapt to varying camera response functions.
% %Dropping exposure normalization causes the largest performance drop, since unaligned brightness across views confuses the fusion.
% %Disabling cross-view attention introduces ghosting on flat surfaces like walls.
% %Finally, omitting upsampling preserves overall structure but loses fine details, producing notably blurry outputs.
% As shown in Tab.~\ref{tab:hdr_syn}~(b) and Fig.~\ref{fig:ablation}, each component plays a critical role. 
% Removing the MetaNet makes training unstable, as the model cannot adapt to varying camera response functions. 
% Eliminating exposure normalization leads to the largest performance degradation, since inconsistent brightness across views disrupts feature fusion. 
% Disabling cross-view attention introduces ghosting artifacts on smooth surfaces such as walls. 
% Finally, removing the upsampling module preserves coarse structure but loses fine details, resulting in blurry outputs.

\subsection{Ablation Study}

\noindent \textbf{Baseline.}
We show that na\"ive extensions of existing pipelines fail.
In Tab.~\ref{tab: fair baseline} (a), we add exposure-aware tokens to AnySplat~\cite{jiang2025anysplat} but it performs limited: \textit{its encoder is geometry-dominated, not color; color is only introduced via skip connection, which lacks cross-view interaction to resolve exposure inconsistency.}
Our key insight is that the frozen geometry encoder already captures reliable cross-view correspondences under large exposure gaps, and its attention maps can be reused for HDR appearance fusion, this has not been explored before.

\begin{table}[t]
\centering
\caption{
(a) Comparison against an exposure-aware AnySplat (HDR) baseline.
(b) Effect of one-time real-scene finetuning.
}
\resizebox{\textwidth}{!}{
\begin{tabular}{cc}

\renewcommand{\arraystretch}{1}
\begin{tabular}{ccccccc}
\multicolumn{7}{c}{\fontsize{11}{11}\selectfont(a) Compared with AnySplat (HDR) baseline.} \\
\toprule
\multirow{2}{*}{HDR-NeRF Real} & \multicolumn{3}{c}{4 Views} & \multicolumn{3}{c}{18 Views} \\
\cmidrule(lr){2-4} \cmidrule(lr){5-7}
 & PSNR$\uparrow$ & SSIM$\uparrow$ & LPIPS$\downarrow$ & PSNR$\uparrow$ & SSIM$\uparrow$ & LPIPS$\downarrow$ \\
\hline
AnySplat (HDR) & 14.59 & 0.571 & 0.419 & 15.65 & 0.629 & 0.377 \\
InstantHDR & \textbf{18.44} & \textbf{0.721} & \textbf{0.269} & \textbf{19.48} & \textbf{0.745} & \textbf{0.257} \\
\bottomrule
\end{tabular}

& 

\renewcommand{\arraystretch}{1}
\begin{tabular}{ccccccc}
\multicolumn{7}{c}{\fontsize{11}{11}\selectfont(b) One-time real-scene finetuning.} \\
\toprule
\multirow{2}{*}{HDR-NeRF Real} & \multicolumn{3}{c}{4 Views} & \multicolumn{3}{c}{18 Views} \\
\cmidrule(lr){2-4} \cmidrule(lr){5-7}
 & PSNR$\uparrow$ & SSIM$\uparrow$ & LPIPS$\downarrow$ & PSNR$\uparrow$ & SSIM$\uparrow$ & LPIPS$\downarrow$ \\
\hline
w/o finetuning & 17.27 & 0.704 & 0.298 & 18.08 & 0.715 & 0.271 \\
w finetuning & 18.44 & 0.721 & 0.269 & 19.48 & 0.745 & 0.257 \\
\bottomrule
\end{tabular}

\label{tab: fair baseline}
\end{tabular}
}
\end{table}

\begin{figure}[t]
\centering
\includegraphics[height=0.125\columnwidth]{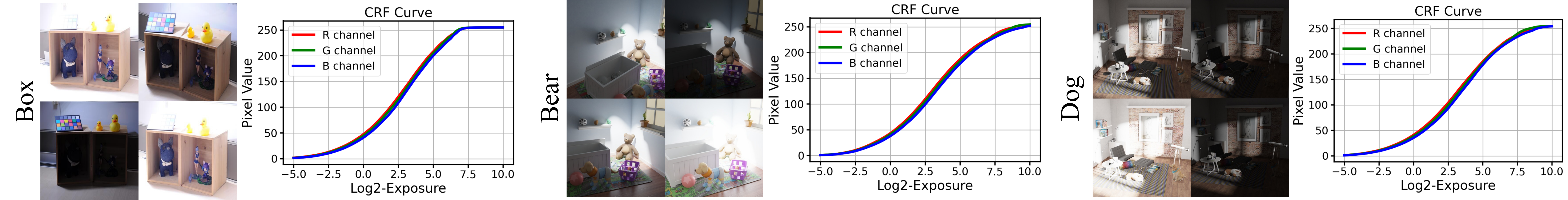}
\caption{
\textbf{CRF visualization.} Learned CRF curves for three scenes, all monotonic.
}
\label{fig: crf}
\end{figure}

\noindent \textbf{Component ablation.}
Tab.~\ref{tab:hdr_syn}~(b) and Fig.~\ref{fig:ablation} confirm each component is critical.
Removing the MetaNet destabilizes training, as the model cannot adapt to varying camera response functions.
Eliminating exposure normalization causes the largest degradation, since inconsistent brightness across views disrupts feature fusion.
Disabling cross-view attention introduces ghosting on smooth surfaces such as walls, confirming the value of reusing the geometry encoder's correspondences.
Removing the upsampling module preserves coarse structure but loses fine details, yielding blurry outputs.

\noindent \textbf{CRF curves.}
To further illustrate the MetaNet ablated above, we show that it automatically learns monotonic, scene-specific CRF curves through large-scale pretraining.
As shown in Fig.~\ref{fig: crf}, the Box scene saturates earlier while others span a wider dynamic range, all strictly monotonic.

\noindent \textbf{Real-scene finetuning.}
Since the pretraining dataset is synthetic, we perform a \emph{one-time finetuning} on 4 HDR-Plenoxels real scenes to bridge the synthetic-to-real gap---not per-scene optimization and no test-data leaking---and apply this model to all real scenes without further tuning.
As shown in Tab.~\ref{tab: fair baseline}~(b), zero-shot (no finetuning) results still generalize well, while finetuning adds $\sim$1.28\,dB.

\section{Conclusion}
\label{sec:conclusion}
We present InstantHDR, the first feed-forward framework for HDR novel view synthesis from uncalibrated multi-exposure LDR images.
By leveraging geometry-guided cross-view attention for exposure-robust appearance fusion and a meta-network for scene-adaptive tone mapping, InstantHDR initializes HDR scenes in few seconds.
We also introduce HDR-Pretrain, a 168-scene synthetic dataset to address the data scarcity for feed-forward HDR pretraining.
Experiments show that, with post-optimization, InstantHDR reaches quality comparable to optimization-based methods at orders-of-magnitude-faster speed.
We hope this InstantHDR can inspire more ideas on real-time 3D HDR reconstruction.

\noindent \textbf{Limitation.}
Our InstantHDR focuses on HDR appearance modeling and relies on feed-forward backbones for pose estimation.
On noise-free dense synthetic scenes, our feed-forward pose estimator is less accurate than COLMAP, which limits reconstruction quality—a limitation shared by all feed-forward methods.

\bibliographystyle{splncs04}
\bibliography{main}

@String(CVPR  = {IEEE Conf. Comput. Vis. Pattern Recog.})

@String(ICCV  = {Int. Conf. Comput. Vis.})

@String(ECCV  = {Eur. Conf. Comput. Vis.})

@String(NeurIPS = {Adv. Neural Inform. Process. Syst.})

@String(ICLR  = {Int. Conf. Learn. Represent.})

@String(BMVC  = {Brit. Mach. Vis. Conf.})

@String(CVPRW = {IEEE Conf. Comput. Vis. Pattern Recog. Worksh.})

@String(AAAI  = {AAAI})

@String(TOG   = {ACM Trans. Graph.})

@String(ACMMM = {ACM Int. Conf. Multimedia})

@String(CVPR  = {CVPR})

@String(ICCV  = {ICCV})

@String(ECCV  = {ECCV})

@String(NeurIPS = {NeurIPS})

@String(ICLR  = {ICLR})

@String(BMVC  =	{BMVC})

@String(CVPRW = {CVPRW})

@String(TOG   = {ACM TOG})

@String(ACMMM = {ACM MM})

@article{mildenhall2021nerf,
  author={Mildenhall, Ben and Srinivasan, Pratul P and Tancik, Matthew and Barron, Jonathan T and Ramamoorthi, Ravi and Ng, Ren},
  title={Nerf: Representing scenes as neural radiance fields for view synthesis},
  journal={Communications of the ACM},
  volume={65},
  number={1},
  pages={99--106},
  year={2021},
}

@inproceedings{yang2020surfelgan,
  author={Yang, Zhenpei and Chai, Yuning and Anguelov, Dragomir and Zhou, Yin and Sun, Pei and Erhan, Dumitru and Rafferty, Sean and Kretzschmar, Henrik},
  title={Surfelgan: Synthesizing realistic sensor data for autonomous driving},
  booktitle=CVPR,
  pages={11118--11127},
  year={2020}
}

@inproceedings{huang2023neural,
  author={Huang, Shengyu and Gojcic, Zan and Wang, Zian and Williams, Francis and Kasten, Yoni and Fidler, Sanja and Schindler, Konrad and Litany, Or},
  title={Neural lidar fields for novel view synthesis},
  booktitle=ICCV,
  pages={18236--18246},
  year={2023}
}

@inproceedings{wang2023sparsenerf,
  author={Wang, Guangcong and Chen, Zhaoxi and Loy, Chen Change and Liu, Ziwei},
  title={Sparsenerf: Distilling depth ranking for few-shot novel view synthesis},
  booktitle=ICCV,
  pages={9065--9076},
  year={2023}
}

@inproceedings{tancik2022block,
  author={Tancik, Matthew and Casser, Vincent and Yan, Xinchen and Pradhan, Sabeek and Mildenhall, Ben and Srinivasan, Pratul P and Barron, Jonathan T and Kretzschmar, Henrik},
  title={Block-nerf: Scalable large scene neural view synthesis},
  booktitle=CVPR,
  pages={8248--8258},
  year={2022}
}

@inproceedings{liu2021editing,
  title={Editing conditional radiance fields},
  author={Liu, Steven and Zhang, Xiuming and Zhang, Zhoutong and Zhang, Richard and Zhu, Jun-Yan and Russell, Bryan},
  booktitle=ICCV,
  pages={5773--5783},
  year={2021}
}

@article{singh24_hdrsplat,
  author    = {Singh, Shreyas and Garg, Aryan and Mitra, Kaushik},
  title     = {HDRSplat: Gaussian Splatting for High Dynamic Range 3D Scene Reconstruction from Raw Images},
  journal   = {BMVC},
  year      = {2024},
}

@article{gong2025casual3dhdr,
  title={Casual3DHDR: Deblurring High Dynamic Range 3D Gaussian Splatting from Casually Captured Videos},
  author={Gong, Shucheng and Zhao, Lingzhe and Li, Wenpu and Xie, Hong and Zhang, Yin and Zhao, Shiyu and Liu, Peidong},
  journal={ACMMM},
  year={2025}
}

@inproceedings{li2025sehdr,
  title={SEHDR: Single-Exposure HDR Novel View Synthesis via 3D Gaussian Bracketing},
  author={Li, Yiyu and Wang, Haoyuan and Xu, Ke and Hancke, Gerhard Petrus and Lau, Rynson WH},
  booktitle={Proceedings of the IEEE/CVF International Conference on Computer Vision},
  pages={26045--26054},
  year={2025}
}

@inproceedings{exposure_fusion,
  title={Exposure fusion},
  author={Mertens, Tom and Kautz, Jan and Van Reeth, Frank},
  booktitle={Pacific Conference on Computer Graphics and Applications},
  year={2007}
}

@inproceedings{malik,
  title={Recovering high dynamic range radiance maps from photographs},
  author={Debevec, Paul E and Malik, Jitendra},
  booktitle={SIGGRAPH},
  year={1997}
}

@inproceedings{ward2008high,
  title={High dynamic range imaging \& image-based lighting},
  author={Ward, Greg and Reinhard, Erik and Debevec, Paul},
  booktitle={SIGGRAPH},
  year={2008}
}

@article{hdr_20,
  title={Fast and robust high dynamic range image generation with camera and object movement},
  author={Grosch, Thorsten and others},
  journal={Vision, Modeling and Visualization, RWTH Aachen},
  year={2006}
}

@article{hdr_24,
  title={Automatic high-dynamic range image generation for dynamic scenes},
  author={Jacobs, Katrien and Loscos, Celine and Ward, Greg},
  journal={IEEE Computer Graphics and Applications},
  year={2008}
}

@article{hdr_26,
  title={Deep high dynamic range imaging of dynamic scenes.},
  author={Kalantari, Nima Khademi and Ramamoorthi, Ravi and others},
  journal={ACM ToG},
  year={2017}
}

@inproceedings{hdr_58,
  title={The state of the art in HDR deghosting: A survey and evaluation},
  author={Tursun, Okan Tarhan and Aky{\"u}z, Ahmet O{\u{g}}uz and Erdem, Aykut and Erdem, Erkut},
  booktitle={Computer Graphics Forum},
  year={2015}
}

@article{hdr_63,
  title={Robust artifact-free high dynamic range imaging of dynamic scenes},
  author={Yan, Qingsen and Zhu, Yu and Zhang, Yanning},
  journal={Multimedia Tools and Applications},
  year={2019}
}

@article{hdr_cnn_1,
  title={HDR image reconstruction from a single exposure using deep CNNs},
  author={Eilertsen, Gabriel and Kronander, Joel and Denes, Gyorgy and Mantiuk, Rafa{\l} K and Unger, Jonas},
  journal={ACM TOG},
  year={2017}
}

@inproceedings{hdr_cnn_2,
  title={Fhdr: Hdr image reconstruction from a single ldr image using feedback network},
  author={Khan, Zeeshan and Khanna, Mukul and Raman, Shanmuganathan},
  booktitle={IEEE Global Conference on Signal and Information Processing},
  year={2019}
}

@inproceedings{hdr_cnn_3,
  title={End-to-end differentiable learning to hdr image synthesis for multi-exposure images},
  author={Kim, Junghee and Lee, Siyeong and Kang, Suk-Ju},
  booktitle={AAAI},
  year={2021}
}

@inproceedings{hdr_trans_1,
  title={Improving dynamic hdr imaging with fusion transformer},
  author={Chen, Rufeng and Zheng, Bolun and Zhang, Hua and Chen, Quan and Yan, Chenggang and Slabaugh, Gregory and Yuan, Shanxin},
  booktitle={AAAI},
  year={2023}
}

@inproceedings{hdr_trans_2,
  title={Ghost-free high dynamic range imaging with context-aware transformer},
  author={Liu, Zhen and Wang, Yinglong and Zeng, Bing and Liu, Shuaicheng},
  booktitle={ECCV},
  year={2022}
}

@inproceedings{neural_gaffer,
      title={Neural Gaffer: Relighting Any Object via Diffusion}, 
      author={Haian Jin and Yuan Li and Fujun Luan and Yuanbo Xiangli and Sai Bi and Kai Zhang and Zexiang Xu and Jin Sun and Noah Snavely},
      booktitle={NeurIPS},
      year={2024},
}

@inproceedings{fei2023generative,
  title={Generative Diffusion Prior for Unified Image Restoration and Enhancement},
  author={Fei, Ben and Lyu, Zhaoyang and Pan, Liang and Zhang, Junzhe and Yang, Weidong and Luo, Tianyue and Zhang, Bo and Dai, Bo},
  booktitle={CVPR},
  year={2023}
}

@inproceedings{mildenhall2022nerf,
  title={Nerf in the dark: High dynamic range view synthesis from noisy raw images},
  author={Mildenhall, Ben and Hedman, Peter and Martin-Brualla, Ricardo and Srinivasan, Pratul P and Barron, Jonathan T},
  booktitle={CVPR},
  year={2022}
}

@article{nazarczuk2022self,
  title={Self-supervised HDR Imaging from Motion and Exposure Cues},
  author={Nazarczuk, Michal and Catley-Chandar, Sibi and Leonardis, Ales and Pellitero, Eduardo P{\'e}rez},
  journal={arXiv preprint arXiv:2203.12311},
  year={2022}
}

@inproceedings{SMAE,
  title={SMAE: Few-shot Learning for HDR Deghosting with Saturation-Aware Masked Autoencoders},
  author={Yan, Qingsen and Zhang, Song and Chen, Weiye and Tang, Hao and Zhu, Yu and Sun, Jinqiu and Van Gool, Luc and Zhang, Yanning},
  booktitle={CVPR},
  year={2023}
}

@inproceedings{2021labeled,
  title={Labeled from unlabeled: Exploiting unlabeled data for few-shot deep hdr deghosting},
  author={Prabhakar, K Ram and Senthil, Gowtham and Agrawal, Susmit and Babu, R Venkatesh and Gorthi, Rama Krishna Sai S},
  booktitle={CVPR},
  year={2021}
}

@inproceedings{magic,
      title={MaGIC: Multi-modality Guided Image Completion},
      author={Yu, Yongsheng and Wang, Hao and Luo, Tiejian and Fan, Heng and Zhang, Libo},
      booktitle={ICLR},
      year={2024}
}

@inproceedings{promptfix,
  title={PromptFix: You Prompt and We Fix the Photo},
  author={Yu, Yongsheng and Zeng, Ziyun and Hua, Hang and Fu, Jianlong and Luo, Jiebo},
  booktitle={NeurIPS},
  year={2024}
}

@inproceedings{hdr_trans_3,
  title={Selective transhdr: Transformer-based selective hdr imaging using ghost region mask},
  author={Song, Jou Won and Park, Ye-In and Kong, Kyeongbo and Kwak, Jaeho and Kang, Suk-Ju},
  booktitle={ECCV},
  year={2022}
}

@inproceedings{hdr_cnn_4,
  title={ADNet: Attention-guided deformable convolutional network for high dynamic range imaging},
  author={Liu, Zhen and Lin, Wenjie and Li, Xinpeng and Rao, Qing and Jiang, Ting and Han, Mingyan and Fan, Haoqiang and Sun, Jian and Liu, Shuaicheng},
  booktitle={CVPRW},
  year={2021}
}

@inproceedings{zhang2018unreasonable,
  title={The unreasonable effectiveness of deep features as a perceptual metric},
  author={Zhang, Richard and Isola, Phillip and Efros, Alexei A and Shechtman, Eli and Wang, Oliver},
  booktitle=CVPR,
  pages={586--595},
  year={2018}
}

@inproceedings{khanna2024habitat,
  title={Habitat synthetic scenes dataset (hssd-200): An analysis of 3d scene scale and realism tradeoffs for objectgoal navigation},
  author={Khanna, Mukul and Mao, Yongsen and Jiang, Hanxiao and Haresh, Sanjay and Shacklett, Brennan and Batra, Dhruv and Clegg, Alexander and Undersander, Eric and Chang, Angel X and Savva, Manolis},
  booktitle=CVPR,
  pages={16384--16393},
  year={2024}
}

@article{dastjerdi2024pandora,
  title={PanDORA: Casual HDR Radiance Acquisition for Indoor Scenes},
  author={Dastjerdi, Mohammad Reza Karimi and Tanguay-Gaudreau, Dominique and Fortier-Chouinard, Fr{\'e}d{\'e}ric and Hold-Geoffroy, Yannick and Demers, Claude and Kalantari, Nima and Lalonde, Jean-Fran{\c{c}}ois},
  journal={arXiv preprint arXiv:2407.06150},
  year={2024}
}

@inproceedings{lu2024pano,
  title={Pano-NeRF: Synthesizing high dynamic range novel views with geometry from sparse low dynamic range panoramic images},
  author={Lu, Zhan and Zheng, Qian and Shi, Boxin and Jiang, Xudong},
  booktitle=AAAI,
  pages={3927--3935},
  year={2024}
}

@inproceedings{ranftl2021vision,
  title={Vision transformers for dense prediction},
  author={Ranftl, Ren{\'e} and Bochkovskiy, Alexey and Koltun, Vladlen},
  booktitle=ICCV,
  pages={12179--12188},
  year={2021}
}

@inproceedings{liu2025mono4dgs,
  title={Mono4DGS-HDR: High Dynamic Range 4D Gaussian Splatting from Alternating-exposure Monocular Videos},
  author={Liu, Jinfeng and Kong, Lingtong and Zhou, Mi and Chen, Jinwen and Xu, Dan},
  booktitle={ICLR},
  year={2026}
}

@inproceedings{bolduc2025GaSLight,
	title={GaSLight: Gaussian Splats for Spatially-Varying Lighting in HDR},
	author={Bolduc, Christophe and Hold-Geoffroy, Yannick and Shu, Zhixin and Lalonde, Jean-Fran{\c{c}}ois},
	booktitle = ICCV,
	year={2025}
}

@inproceedings{perez2018film,
  title={Film: Visual reasoning with a general conditioning layer},
  author={Perez, Ethan and Strub, Florian and De Vries, Harm and Dumoulin, Vincent and Courville, Aaron},
  booktitle=AAAI,
  year={2018}
}

@inproceedings{SelfHDR,
    title={Self-Supervised High Dynamic Range Imaging with Multi-Exposure Images in Dynamic Scenes},
    author={Zhang, Zhilu and Wang, Haoyu and Liu, Shuai and Wang, Xiaotao and Lei, Lei and Zuo, Wangmeng},
    booktitle={ICLR},
    year={2024}
}

@article{oquab2023dinov2,
  title={Dinov2: Learning robust visual features without supervision},
  author={Oquab, Maxime and Darcet, Timoth{\'e}e and Moutakanni, Th{\'e}o and Vo, Huy and Szafraniec, Marc and Khalidov, Vasil and Fernandez, Pierre and Haziza, Daniel and Massa, Francisco and El-Nouby, Alaaeldin and others},
  journal={arXiv preprint arXiv:2304.07193},
  year={2023}
}

@article{sun2022ide,
  author={Sun, Jingxiang and Wang, Xuan and Shi, Yichun and Wang, Lizhen and Wang, Jue and Liu, Yebin},
  title={Ide-3d: Interactive disentangled editing for high-resolution 3d-aware portrait synthesis},
  journal=TOG,
  volume={41},
  number={6},
  pages={1--10},
  year={2022},
}

@inproceedings{yuan2022nerf,
  author={Yuan, Yu-Jie and Sun, Yang-Tian and Lai, Yu-Kun and Ma, Yuewen and Jia, Rongfei and Gao, Lin},
  title={Nerf-editing: geometry editing of neural radiance fields},
  booktitle=CVPR,
  pages={18353--18364},
  year={2022}
}

@article{kobayashi2022decomposing,
  author={Kobayashi, Sosuke and Matsumoto, Eiichi and Sitzmann, Vincent},
  title={Decomposing nerf for editing via feature field distillation},
  journal=NeurIPS,
  volume={35},
  pages={23311--23330},
  year={2022}
}

@article{liu2021neural,
  author={Liu, Lingjie and Habermann, Marc and Rudnev, Viktor and Sarkar, Kripasindhu and Gu, Jiatao and Theobalt, Christian},
  title={Neural actor: Neural free-view synthesis of human actors with pose control},
  journal=TOG,
  volume={40},
  number={6},
  pages={1--16},
  year={2021},
}

@inproceedings{hu2021egorenderer,
  author={Hu, Tao and Sarkar, Kripasindhu and Liu, Lingjie and Zwicker, Matthias and Theobalt, Christian},
  title={Egorenderer: Rendering human avatars from egocentric camera images},
  booktitle=ICCV,
  pages={14528--14538},
  year={2021}
}

@inproceedings{zheng2023ilsh,
  author={Zheng, Jiali and Jang, Youngkyoon and Papaioannou, Athanasios and Kampouris, Christos and Potamias, Rolandos Alexandros and Papantoniou, Foivos Paraperas and Galanakis, Efstathios and Leonardis, Ale{\v{s}} and Zafeiriou, Stefanos},
  title={Ilsh: The imperial light-stage head dataset for human head view synthesis},
  booktitle=ICCV,
  pages={1112--1120},
  year={2023}
}

@inproceedings{zheng2022structured,
  author={Zheng, Zerong and Huang, Han and Yu, Tao and Zhang, Hongwen and Guo, Yandong and Liu, Yebin},
  title={Structured local radiance fields for human avatar modeling},
  booktitle=CVPR,
  pages={15893--15903},
  year={2022}
}

@inproceedings{huang2022hdr,
  author={Huang, Xin and Zhang, Qi and Feng, Ying and Li, Hongdong and Wang, Xuan and Wang, Qing},
  title={Hdr-nerf: High dynamic range neural radiance fields},
  booktitle=CVPR,
  pages={18398--18408},
  year={2022}
}

@article{cai2024hdr,
  author={Cai, Yuanhao and Xiao, Zihao and Liang, Yixun and Qin, Minghan and Zhang, Yulun and Yang, Xiaokang and Liu, Yaoyao and Yuille, Alan L},
  title={Hdr-gs: Efficient high dynamic range novel view synthesis at 1000x speed via gaussian splatting},
  journal=NeurIPS,
  volume={37},
  pages={68453--68471},
  year={2024}
}

@inproceedings{liu2025gausshdr,
  author={Liu, Jinfeng and Kong, Lingtong and Li, Bo and Xu, Dan},
  title={GaussHDR: High Dynamic Range Gaussian Splatting via Learning Unified 3D and 2D Local Tone Mapping},
  booktitle=CVPR,
  pages={5991--6000},
  year={2025}
}

@inproceedings{wang2025vggt,
  author={Wang, Jianyuan and Chen, Minghao and Karaev, Nikita and Vedaldi, Andrea and Rupprecht, Christian and Novotny, David},
  title={Vggt: Visual geometry grounded transformer},
  booktitle=CVPR,
  pages={5294--5306},
  year={2025}
}

@article{jiang2025anysplat,
  author={Jiang, Lihan and Mao, Yucheng and Xu, Linning and Lu, Tao and Ren, Kerui and Jin, Yichen and Xu, Xudong and Yu, Mulin and Pang, Jiangmiao and Zhao, Feng and others},
  title={Anysplat: Feed-forward 3d gaussian splatting from unconstrained views},
  journal=TOG,
  volume={44},
  number={6},
  pages={1--16},
  year={2025},
}

@article{3dgs,
	title={3D Gaussian Splatting for Real-Time Radiance Field Rendering},
	author={Kerbl, Bernhard and Kopanas, Georgios and Leimk{\"u}hler, Thomas and Drettakis, George},
	journal={ACM Transactions on Graphics},
	year={2023}
}

@article{dynamic1,
	title={4d gaussian splatting for real-time dynamic scene rendering},
	author={Wu, Guanjun and Yi, Taoran and Fang, Jiemin and Xie, Lingxi and Zhang, Xiaopeng and Wei, Wei and Liu, Wenyu and Tian, Qi and Wang, Xinggang},
	journal={arXiv preprint arXiv:2310.08528},
	year={2023}
}

@article{dynamic2,
	title={Dynamic 3d gaussians: Tracking by persistent dynamic view synthesis},
	author={Luiten, Jonathon and Kopanas, Georgios and Leibe, Bastian and Ramanan, Deva},
	journal={arXiv preprint arXiv:2308.09713},
	year={2023}
}

@article{dynamic3,
	title={Real-time photorealistic dynamic scene representation and rendering with 4d gaussian splatting},
	author={Yang, Zeyu and Yang, Hongye and Pan, Zijie and Zhu, Xiatian and Zhang, Li},
	journal={arXiv preprint arXiv:2310.10642},
	year={2023}
}

@article{slam3,
	title={Splatam: Splat, track \& map 3d gaussians for dense rgb-d slam},
	author={Keetha, Nikhil and Karhade, Jay and Jatavallabhula, Krishna Murthy and Yang, Gengshan and Scherer, Sebastian and Ramanan, Deva and Luiten, Jonathon},
	journal={arXiv preprint arXiv:2312.02126},
	year={2023}
}

@article{slam4,
	title={Gaussian-SLAM: Photo-realistic Dense SLAM with Gaussian Splatting},
	author={Yugay, Vladimir and Li, Yue and Gevers, Theo and Oswald, Martin R},
	journal={arXiv preprint arXiv:2312.10070},
	year={2023}
}

@article{slam1,
	title={GS-SLAM: Dense Visual SLAM with 3D Gaussian Splatting},
	author={Yan, Chi and Qu, Delin and Wang, Dong and Xu, Dan and Wang, Zhigang and Zhao, Bin and Li, Xuelong},
	journal={arXiv preprint arXiv:2311.11700},
	year={2023}
}

@article{slam2,
	title={Gaussian Splatting SLAM},
	author={Matsuki, Hidenobu and Murai, Riku and Kelly, Paul HJ and Davison, Andrew J},
	journal={arXiv preprint arXiv:2312.06741},
	year={2023}
}

@article{InverseRendering2,
	title={GS-IR: 3D Gaussian Splatting for Inverse Rendering},
	author={Liang, Zhihao and Zhang, Qi and Feng, Ying and Shan, Ying and Jia, Kui},
	journal={arXiv preprint arXiv:2311.16473},
	year={2023}
}

@article{InverseRendering3,
	title={Physgaussian: Physics-integrated 3d gaussians for generative dynamics},
	author={Xie, Tianyi and Zong, Zeshun and Qiu, Yuxin and Li, Xuan and Feng, Yutao and Yang, Yin and Jiang, Chenfanfu},
	journal={arXiv preprint arXiv:2311.12198},
	year={2023}
}

@article{InverseRendering1,
	title={GaussianShader: 3D Gaussian Splatting with Shading Functions for Reflective Surfaces},
	author={Jiang, Yingwenqi and Tu, Jiadong and Liu, Yuan and Gao, Xifeng and Long, Xiaoxiao and Wang, Wenping and Ma, Yuexin},
	journal={arXiv preprint arXiv:2311.17977},
	year={2023}
}

@article{humangaussian,
	title={Humangaussian: Text-driven 3d human generation with gaussian splatting},
	author={Liu, Xian and Zhan, Xiaohang and Tang, Jiaxiang and Shan, Ying and Zeng, Gang and Lin, Dahua and Liu, Xihui and Liu, Ziwei},
	journal={arXiv preprint arXiv:2311.17061},
	year={2023}
}

@article{gaussiandreamer,
	title={Gaussiandreamer: Fast generation from text to 3d gaussian splatting with point cloud priors},
	author={Yi, Taoran and Fang, Jiemin and Wu, Guanjun and Xie, Lingxi and Zhang, Xiaopeng and Liu, Wenyu and Tian, Qi and Wang, Xinggang},
	journal={arXiv preprint arXiv:2310.08529},
	year={2023}
}

@article{luciddreamer,
	title={LucidDreamer: Towards High-Fidelity Text-to-3D Generation via Interval Score Matching},
	author={Liang, Yixun and Yang, Xin and Lin, Jiantao and Li, Haodong and Xu, Xiaogang and Chen, Yingcong},
	journal={arXiv preprint arXiv:2311.11284},
	year={2023}
}

@inproceedings{x_gaussian,
  title={Radiative Gaussian Splatting for Efficient X-ray Novel View Synthesis},
  author={Cai, Yuanhao and Liang, Yixun and Wang, Jiahao and Wang, Angtian and Zhang, Yulun and Yang, Xiaokang and Zhou, Zongwei and Yuille, Alan},
  booktitle={ECCV},
  year={2024}
}

@inproceedings{wang2024dust3r,
  title={Dust3r: Geometric 3d vision made easy},
  author={Wang, Shuzhe and Leroy, Vincent and Cabon, Yohann and Chidlovskii, Boris and Revaud, Jerome},
  booktitle=CVPR,
  pages={20697--20709},
  year={2024}
}

@inproceedings{mast3r,
  title={Grounding image matching in 3d with mast3r},
  author={Leroy, Vincent and Cabon, Yohann and Revaud, J{\'e}r{\^o}me},
  booktitle=ECCV,
  pages={71--91},
  year={2024},
  organization={Springer}
}

@article{wang20243d,
  title={3d reconstruction with spatial memory},
  author={Wang, Hengyi and Agapito, Lourdes},
  journal={arXiv preprint arXiv:2408.16061},
  year={2024}
}

@inproceedings{liu2025slam3r,
  title={Slam3r: Real-time dense scene reconstruction from monocular rgb videos},
  author={Liu, Yuzheng and Dong, Siyan and Wang, Shuzhe and Yin, Yingda and Yang, Yanchao and Fan, Qingnan and Chen, Baoquan},
  booktitle=CVPR,
  pages={16651--16662},
  year={2025}
}

@inproceedings{murai2025mast3r,
  title={MASt3R-SLAM: Real-time dense SLAM with 3D reconstruction priors},
  author={Murai, Riku and Dexheimer, Eric and Davison, Andrew J},
  booktitle=CVPR,
  pages={16695--16705},
  year={2025}
}

@article{yang2025fast3r,
  title={Fast3R: Towards 3D Reconstruction of 1000+ Images in One Forward Pass},
  author={Yang, Jianing and Sax, Alexander and Liang, Kevin J and Henaff, Mikael and Tang, Hao and Cao, Ang and Chai, Joyce and Meier, Franziska and Feiszli, Matt},
  journal={arXiv preprint arXiv:2501.13928},
  year={2025}
}

@inproceedings{tang2025mv,
  title={Mv-dust3r+: Single-stage scene reconstruction from sparse views in 2 seconds},
  author={Tang, Zhenggang and Fan, Yuchen and Wang, Dilin and Xu, Hongyu and Ranjan, Rakesh and Schwing, Alexander and Yan, Zhicheng},
  booktitle=CVPR,
  pages={5283--5293},
  year={2025}
}

@article{wang2023pf,
  title={Pf-lrm: Pose-free large reconstruction model for joint pose and shape prediction},
  author={Wang, Peng and Tan, Hao and Bi, Sai and Xu, Yinghao and Luan, Fujun and Sunkavalli, Kalyan and Wang, Wenping and Xu, Zexiang and Zhang, Kai},
  journal={arXiv preprint arXiv:2311.12024},
  year={2023}
}

@article{hong2024pf3plat,
  title   = {PF3plat: Pose-Free Feed-Forward 3D Gaussian Splatting},
  author  = {Sunghwan Hong and Jaewoo Jung and Heeseong Shin and Jisang Han and Jiaolong Yang and Chong Luo and Seungryong Kim},
  journal = {arXiv preprint arXiv:2410.22128},
  year    = {2024}
}

@article{ye2024no,
  title={No pose, no problem: Surprisingly simple 3d gaussian splats from sparse unposed images},
  author={Ye, Botao and Liu, Sifei and Xu, Haofei and Li, Xueting and Pollefeys, Marc and Yang, Ming-Hsuan and Peng, Songyou},
  journal={arXiv preprint arXiv:2410.24207},
  year={2024}
}

@article{zhang2025flare,
  title={Flare: Feed-forward geometry, appearance and camera estimation from uncalibrated sparse views},
  author={Zhang, Shangzhan and Wang, Jianyuan and Xu, Yinghao and Xue, Nan and Rupprecht, Christian and Zhou, Xiaowei and Shen, Yujun and Wetzstein, Gordon},
  journal={arXiv preprint arXiv:2502.12138},
  year={2025}
}

@article{smart2024splatt3r,
  title={Splatt3r: Zero-shot gaussian splatting from uncalibrated image pairs},
  author={Smart, Brandon and Zheng, Chuanxia and Laina, Iro and Prisacariu, Victor Adrian},
  journal={arXiv preprint arXiv:2408.13912},
  year={2024}
}

@article{chen2024pref3r,
  title={PreF3R: Pose-Free Feed-Forward 3D Gaussian Splatting from Variable-length Image Sequence},
  author={Chen, Zequn and Yang, Jiezhi and Yang, Heng},
  journal={arXiv preprint arXiv:2411.16877},
  year={2024}
}

@article{jiang2023leap,
  title={Leap: Liberate sparse-view 3d modeling from camera poses},
  author={Jiang, Hanwen and Jiang, Zhenyu and Zhao, Yue and Huang, Qixing},
  journal={arXiv preprint arXiv:2310.01410},
  year={2023}
}

@article{wang2025continuous,
  title={Continuous 3D Perception Model with Persistent State},
  author={Wang, Qianqian and Zhang, Yifei and Holynski, Aleksander and Efros, Alexei A and Kanazawa, Angjoo},
  journal={arXiv preprint arXiv:2501.12387},
  year={2025}
}

@inproceedings{r2gs,
  title={R$^2$-Gaussian: Rectifying Radiative Gaussian Splatting for Tomographic Reconstruction},
  author={Zha, Ruyi and Lin, Tao Jun and Cai, Yuanhao and Cao, Jiwen and Zhang, Yanhao and Li, Hongdong},
  booktitle={NeurIPS},
  year={2024}
}

@inproceedings{gauhuman,
  title={Gauhuman: Articulated gaussian splatting from monocular human videos},
  author={Hu, Shoukang and Hu, Tao and Liu, Ziwei},
  booktitle={CVPR},
  pages={20418--20431},
  year={2024}
}

@article{dreamgaussian,
	title={Dreamgaussian: Generative gaussian splatting for efficient 3d content creation},
	author={Tang, Jiaxiang and Ren, Jiawei and Zhou, Hang and Liu, Ziwei and Zeng, Gang},
	journal={arXiv preprint arXiv:2309.16653},
	year={2023}
}

@inproceedings{cui2025luminance,
  title={Luminance-gs: Adapting 3d gaussian splatting to challenging lighting conditions with view-adaptive curve adjustment},
  author={Cui, Ziteng and Chu, Xuangeng and Harada, Tatsuya},
  booktitle=CVPR,
  pages={26472--26482},
  year={2025}
}

@inproceedings{li2024chaos,
  title={From chaos to clarity: 3DGS in the dark},
  author={Li, Zhihao and Wang, Yufei and Kot, Alex and Wen, Bihan},
  booktitle={NeurIPS},
  volume={37},
  pages={94971--94992},
  year={2024}
}

@inproceedings{zhou2025lita,
  title={LITA-GS: Illumination-agnostic novel view synthesis via reference-free 3D Gaussian splatting and physical priors},
  author={Zhou, Han and Dong, Wei and Chen, Jun},
  booktitle=CVPR,
  pages={21580--21589},
  year={2025}
}

@inproceedings{wu2024fast,
  title={Fast High Dynamic Range Radiance Fields for Dynamic Scenes},
  author={Wu, Guanjun and Yi, Taoran and Fang, Jiemin and Liu, Wenyu and Wang, Xinggang},
  booktitle={2024 International Conference on 3D Vision (3DV)},
  pages={862--872},
  year={2024},
  organization={IEEE}
}

@article{jin2024lighting,
  title={Lighting every darkness with 3dgs: Fast training and real-time rendering for hdr view synthesis},
  author={Jin, Xin and Jiao, Pengyi and Duan, Zheng-Peng and Yang, Xingchao and Li, Chongyi and Guo, Chun-Le and Ren, Bo},
  journal={NeurIPS},
  volume={37},
  pages={80191--80219},
  year={2024}
}

@article{hugs,
	title={Hugs: Human gaussian splats},
	author={Kocabas, Muhammed and Chang, Jen-Hao Rick and Gabriel, James and Tuzel, Oncel and Ranjan, Anurag},
	journal={arXiv preprint arXiv:2311.17910},
	year={2023}
}

@inproceedings{jun2022hdr,
  author={Jun-Seong, Kim and Yu-Ji, Kim and Ye-Bin, Moon and Oh, Tae-Hyun},
  title={Hdr-plenoxels: Self-calibrating high dynamic range radiance fields},
  booktitle=ECCV,
  pages={384--401},
  year={2022},
  organization={Springer}
}

@inproceedings{zou2023rawhdr,
  title={RawHDR: High Dynamic Range Image Reconstruction from a Single Raw Image},
  author={Zou, Yunhao and Yan, Chenggang and Fu, Ying},
  booktitle=ICCV,
  year={2023}
}

@inproceedings{liu2023joint,
  title={Joint HDR Denoising and Fusion: A Real-World Mobile HDR Image Dataset},
  author={Liu, Shuaizheng and Zhang, Xindong and Sun, Lingchen and Liang, Zhetong and Zeng, Hui and Zhang, Lei},
  booktitle=CVPR,
  year={2023}
}

@inproceedings{shu2024real,
  title={Towards Real-World HDR Video Reconstruction: A Large-Scale Benchmark Dataset and A Two-Stage Alignment Network},
  author={Shu, Yong and Shen, Liquan and Hu, Xiangyu and Li, Mengyao and Zhou, Zihao},
  booktitle=CVPR,
  year={2024}
}

@inproceedings{xu2024hdrflow,
  title={HDRFlow: Real-Time HDR Video Reconstruction with Large Motions},
  author={Xu, Gangwei and Wang, Yujin and Gu, Jinwei and Xue, Tianfan and Yang, Xin},
  booktitle=CVPR,
  year={2024}
}

@inproceedings{li2024dualdn,
  title={DualDn: Dual-Domain Denoising via Differentiable ISP},
  author={Li, Ruikang and Wang, Yujin and Chen, Shiqi and Zhang, Fan and Gu, Jinwei and Xue, Tianfan},
  booktitle=ECCV,
  year={2024}
}

@inproceedings{lee2025ntire,
  title={NTIRE 2025 challenge on efficient burst hdr and restoration: Datasets, methods, and results},
  author={Lee, Sangmin and Park, Eunpil and Canelo, Angel and Park, Hyunhee and Kim, Youngjo and Chun, Hyungju and Jin, Xin and Li, Chongyi and Guo, Chun-Le and Timofte, Radu and others},
  booktitle=CVPRW,
  pages={1002--1017},
  year={2025}
}

@inproceedings{jiang2025learning,
  title={Learning to See in the Extremely Dark},
  author={Jiang, Hai and Guan, Binhao and Liu, Zhen and Liu, Xiaohong and Yu, Jian and Liu, Zheng and Han, Songchen and Liu, Shuaicheng},
  booktitle=ICCV,
  year={2025}
}
\end{document}